\documentclass[letterpaper,12pt]{article}
\usepackage[margin=1in,dvips]{geometry}
\usepackage{graphicx,float,psfrag,amsmath,amsthm,amssymb,nicefrac}
\usepackage[english]{babel}
\usepackage{listings}
\usepackage{soul}
\usepackage{multirow}
\usepackage{algpseudocode}
\usepackage{algorithm,algpseudocode}
\usepackage{natbib}
\usepackage{enumitem}
\usepackage{url,color,booktabs,verbatim}

\title{Learning Representations in Model-Free \\Hierarchical Reinforcement Learning}
\author{Jacob Rafati \& David C. Noelle\\ \texttt{\{jrafatiheravi,dnoelle\}@ucmerced.edu} \\ Electrical Engineering and Computer Science\\Computational Cognitive Neuroscience Laboratory\\
Univeristy of California, Merced\\5200 North Lake Road, Merced, CA, 95343, USA.}

\date{}
\begin{document}

\maketitle

\section*{Abstract}
Common approaches to Reinforcement Learning (RL) are seriously challenged by large-scale applications involving huge state spaces and sparse delayed reward feedback. Hierarchical Reinforcement Learning (HRL) methods attempt to address this scalability issue by learning action selection policies at multiple levels of \emph{temporal abstraction}. Abstraction can be had by identifying a relatively small set of states that are likely to be useful as subgoals, in concert with the learning of corresponding skill policies to achieve those subgoals. Many approaches to \emph{subgoal discovery} in HRL depend on the analysis of a model of the environment, but the need to learn such a model introduces its own problems of scale. Once subgoals are identified, skills may be learned through \emph{intrinsic motivation}, introducing an internal reward signal marking subgoal attainment. In this paper, we present a novel model-free method for subgoal discovery using incremental unsupervised learning over a small memory of the most recent experiences (trajectories) of the agent. When combined with an intrinsic motivation learning mechanism, this method learns both subgoals and skills, based on experiences in the environment. Thus, we offer an original approach to HRL that does not require the acquisition of a model of the environment, suitable for large-scale applications. We demonstrate the efficiency of our method on two RL problems with sparse delayed feedback: a variant of the rooms environment and the first screen of the ATARI 2600 \emph{Montezuma's Revenge} game.

\section{Introduction}
The reinforcement learning problem suffers from serious scaling
issues. Methods such as transfer learning \citep{Ammar2012TransferRL,singh1992transfer,taylor2009transfer}, and Hierarchical Reinforcement Learning (HRL) attempt to address these issues \citep{Barto:2003:HRL,ENC-HRL:Hengst2010,dayan1992feudal,dietterich2000hierarchical}. HRL is an important
computational approach intended to tackle problems of scale by
learning to operate over different levels of \emph{temporal
	abstraction} \citep{Sutton:1999:Option,parr1997reinforcement,Krishnamurthy2016HierarchicalRL,Stolle:2002:LOR}. The acquisition of
hierarchies of reusable skills is one of the distinguishing
characteristics of biological intelligence \citep{Botvinick:2009:HRL,diuk2013divide,Badre:2010:PFC-HRL}, and the learning of such hierarchies is an important open problem in computational reinforcement learning. The development of robust HRL methods will provide a means to acquire relevant knowledge at multiple levels of abstraction, potentially speeding learning and supporting generalization.

A number of approaches to HRL have been suggested. One approach
focuses on action sequences, subpolicies, or ``options'' that appear
repeatedly during the learning of a set of tasks \citep{Sutton:1999:Option,levy2011unified,Fox2017MultiLevelDO,bacon2017option,Stolle:2002:LOR,Bakker04hierarchicalreinforcement}. Such frequently reused subpolicies can be abstracted into skills that can be treated as individual actions at a higher level of abstraction. A somewhat
different approach to temporal abstraction involves identifying a set of states that make for useful \emph{subgoals} \citep{Goel:2003:Subgoal,Simsek:2005:subgoal,McGovern2001Subgoal,Machado:2016:Purposeful}. This introduces a major open problem in HRL: that of
\emph{subgoal discovery}.

A variety of researchers have proposed approaches to identifying
useful subpolicies and reifying them as
skills \citep{Pickett-and-Barto:2002:policy-blocks,Thrun-and-Schwartz:1995:NIPS-Finding-Structure,Mannor:2004:DAR,Stolle:2002:LOR}. For
example, \cite{Sutton:1999:Option} proposed the \emph{options
	framework}, which involves abstractions over the space of actions. At
each step, the agent chooses either a one-step ``primitive'' action or
a ``multi-step'' action policy (an option). Each option defines a
policy over actions (either primitive or other options) and comes to
completion according to a termination condition. Other researchers
have focused on identifying subgoals --- states that are generally
useful to attain --- and learning a collection of skills that allow
the agent to efficiently reach those subgoals. Some approaches to
subgoal discovery maintain the value function in a large look-up
table \citep{Sutton:1999:Option,Goel:2003:Subgoal,Simsek:2005:subgoal,McGovern2001Subgoal},
and most of these methods require building the state transition graph,
providing a model of the environment and the agents possible
interactions with
it \citep{Machado:2017:Laplacian,Simsek:2005:subgoal,Goel:2003:Subgoal,Mannor:2004:DAR,Stolle:2002:LOR}. Formally,
the state transition graph is a directed graph $G=(V,E)$ with a set of
vertices, $V \subseteq \mathcal{S}$ and set of edges $E \subseteq
{\mathcal{A}(\mathcal{S})}$, where $\mathcal{S}$ is the set of states
and $\mathcal{A}(\mathcal{S})$ is the set of allowable actions when in a given state. Since
actions typically modify the state of the agent, each directed edge,
$(s,s') \in E$, indicates an action that takes the agent from state
$s$ to state $s'$. In nondeterministic environments, a probability
distribution over subsequent states, given the current state and an
action, $p(s' | s,a)$, is maintained as part of the model of the
environment. {One method of this kind that was applied to a somewhat
	larger scale task --- the first screen of the ATARI 2600 game
	called \emph{Montezuma's Revenge} --- is that of \cite{Machado:2016:Purposeful}.
	This method constructs the combinatorial transition graph Laplacian
	matrix, and an eigen-decomposition of that matrix produces candidate
	subgoals. While it was shown that some of these candidates make for
	useful subgoals, only heuristic domain-sensitive methods have been
	reported for identifying useful subgoals from the large set of
	candidates (e.g., thousands). Thus, previously} proposed subgoal discovery methods have provided useful insights and
have been demonstrated to improve learning, but there continue to be challenges with regard to
scalability and generalization. Scaling to large state spaces will generally mandate the use of some
form of nonlinear function approximator to encode the value function,
rather than a look-up table. More importantly, as the scale of
a reinforcement learning problem increases, the tractability of
obtaining a good model of the environment, capturing all relevant
state transition probabilities, precipitously decreases.

Once useful subgoals are discovered, an HRL agent should be able to
learn the skills to attain those subgoals through the use of
\emph{intrinsic motivation} --- artificially rewarding the agent for
attaining selected subgoals \citep{Singh:2010:intrinsic-motivation,Vigorito:2010:intrinsic-motivation}. In such systems, knowledge of the current subgoal is needed to estimate future intrinsic reward, resulting in value functions that consider subgoals along with states \citep{vezhnevets2017feudal}. The nature and origin of ``good''
intrinsic reward functions is an open question in reinforcement
learning, however, and a number of approaches have been
proposed. \cite{Singh:2010:intrinsic-motivation}
explored agents with intrinsic reward structures in order to learn
generic options that can apply to a wide variety of tasks. Value
functions have also been generalized to consider goals along with
states \citep{vezhnevets2017feudal}. Such a parameterized universal value
function, $q(s,g,a;w)$, integrates the value functions for multiple
skills into a single function taking the current subgoal,
$g$, as an argument.

Recently, \cite{Kulkarni:2016:Meta-Controller}
proposed a scheme for temporal abstraction that involves
simultaneously learning options and a hierarchical control policy in a
deep reinforcement learning framework. Their approach does not use
separate $Q$-functions for each option, but, instead, treats the
option as an argument. This method lacks a technique for automatic
subgoal discovery, however, forcing the system designer to specify a
set of promising subgoal candidates in advance. The approach proposed
in this paper is inspired by \cite{Kulkarni:2016:Meta-Controller}, which has advantages
in terms of scalability and generalization, but it incorporates
automatic subgoal discovery.

It is important to note that \emph{model-free} HRL, which does not
require a model of the environment, still often requires the learning
of useful internal representations of states. When learning the value
function using a nonlinear function approximator, such as a deep
neural network, relevant features of states must be extracted in order
to support generalization at scale. A number of methods have been
explored for learning such internal representations during model-free
RL \citep{TesauroG:1995:TDGammon,Rafati-Noelle:2017:CCCN,DeepMind:Nature:2015}, and deep model-based HRL \citep{Kulkarni:2016:Meta-Controller,Li2017MBHRL}. However, selecting the right representation is still an open problem \citep{Maillard2011RepRL}.

In this paper, we seek to address major open problems in the
integration of internal representation learning, temporal abstraction,
automatic subgoal discovery, and intrinsic motivation learning, all
within the model-free HRL framework \citep{Rafati-Noelle:2019:AAAI}. We propose and implement efficient and general methods for subgoal discovery using unsupervised learning and anomaly (outlier) detection \citep{,Rafati-Noelle:2019:AAAI-KEG}. These methods do not require information beyond that which is typically collected by the agent during model-free reinforcement learning, such as a small memory of recent experiences {(agent trajectories)}. Our methods are fundamentally constrained in three ways, by design. First, we remain faithful to a model-free reinforcement learning framework, eschewing any approach that requires the learning or use of an environment model. Second, we are devoted to integrating subgoal discovery with intrinsic motivation
learning, and temporal abstraction. Specifically, we conjecture that intrinsic motivation
learning can increase appropriate state space coverage, supporting
more efficient subgoal discovery. Lastly, we focus on subgoal
discovery algorithms that are likely to scale to large reinforcement
learning tasks. The result is a unified model-free HRL algorithm that incorporates the learning of useful internal representations of states, automatic subgoal
discovery, intrinsic motivation learning of skills, and the learning
of subgoal selection by a ``meta-controller''. We
demonstrate the effectiveness of this algorithm by applying it to a
variant of the rooms task (illustrated in Figure \ref{fig:rooms}(a)),
as well as the initial screen of the ATARI 2600 game called
\emph{Montezuma's Revenge} (illustrated in
Figure \ref{fig:montezuma}(a)).

\section{Failure of RL in Tasks with Sparse Feedback}
In an RL problem, the agent should implement a policy, $\pi$, from states, $\mathcal{S}$, to possible actions, $\mathcal{A}$, to maximize its expected return from the environment \citep{RL-Book:Sutton:Barto:2017}. At each cycle of interaction, the agent receives a state, $s$, from the environment, takes an action, $a$, and one time step later, the environment sends a reward, $r \in \mathbb{R}$, and an updated state, $s'$. Each cycle of interaction, $e = (s,a,r,s')$ is called a transition \emph{experience}. The goal is to find an optimal policy that maximizes the expected value of the return, i.e. the cumulative sum of future rewards, $G_t = \sum_{t'=t}^T \gamma^{t'-t} r_{t'+1}$, where $\gamma \in [0,1]$ is a discount factor, and $T$ as a final step. It is often useful to define a parametrized value function $Q(s,a;w)$ to estimate the expected value of the return. Q-learning is a Temporal Difference (model-free RL) algorithm that attempts to find the optimal value function by minimizing the loss function, $L(w)$, which is defined over a recent \emph{experience memory}, $\mathcal{D}$: 
\begin{align}
L(w) \triangleq \mathbb{E}_{(s,a,r,s') \sim \mathcal{D}} \Big[ \big( r + \gamma \max_{a'}Q(s',a';w) - Q(s,a;w) \big)^2\Big].
\label{eq:regular-rl-loss}
\end{align}

Learning representations of the value function is challenging for tasks with sparse, and delayed rewards, since in \eqref{eq:regular-rl-loss}, $r=0$ (or an undiagnostic constant value such as $r=-1$) for most experiences. Even if the agent accidentally visits a rare rewarding state, where $r>0$, the experience replay mechanism often fails to learn the value of those states \citep{DeepMind:Nature:2015}. 

Another major problem in RL is the exploration-exploitation trade-off. Common approaches, such as the $\epsilon$-greedy method, are not sufficiently efficient in exploring the state space to succeed on large-scale complex problems \citep{Bellemare2016UCB,Vigorito:2010:intrinsic-motivation}. As a simple example, consider the task of navigation in
the \emph{4-room environment with a key and a car}, shown in
Figure \ref{grid-world-key-car-env} The agent is rewarded for entering the grid square containing the key, and it is more substantially rewarded for entering the grid square with the car after obtaining the key. The other states are not rewarded. Learning even this simple task is challenging for a reinforcement learning agent.

Our intuition, shared with other researchers, is that hierarchies of
abstraction will be critical for successfully solving problems of this
kind. To be successful, the agent should represent knowledge at
multiple levels of spatial and temporal abstraction. Appropriate abstraction can be had by identifying a relatively small set of states that are likely to be useful as \emph{subgoals} and jointly learning the corresponding skills of achieving those subgoals, using intrinsic motivation. 

\section{Hierarchical Reinforcement Learning}
In Hierarchical Reinforcement Learning, a central goal is to allow learning to happen simultaneously on several levels of abstraction. As a simple illustration of the problem, consider the task of navigation in the \emph{4-room environment with a key and a car}. 

\begin{figure}[hbt!]
	\centering
	\includegraphics[width=.6\textwidth]{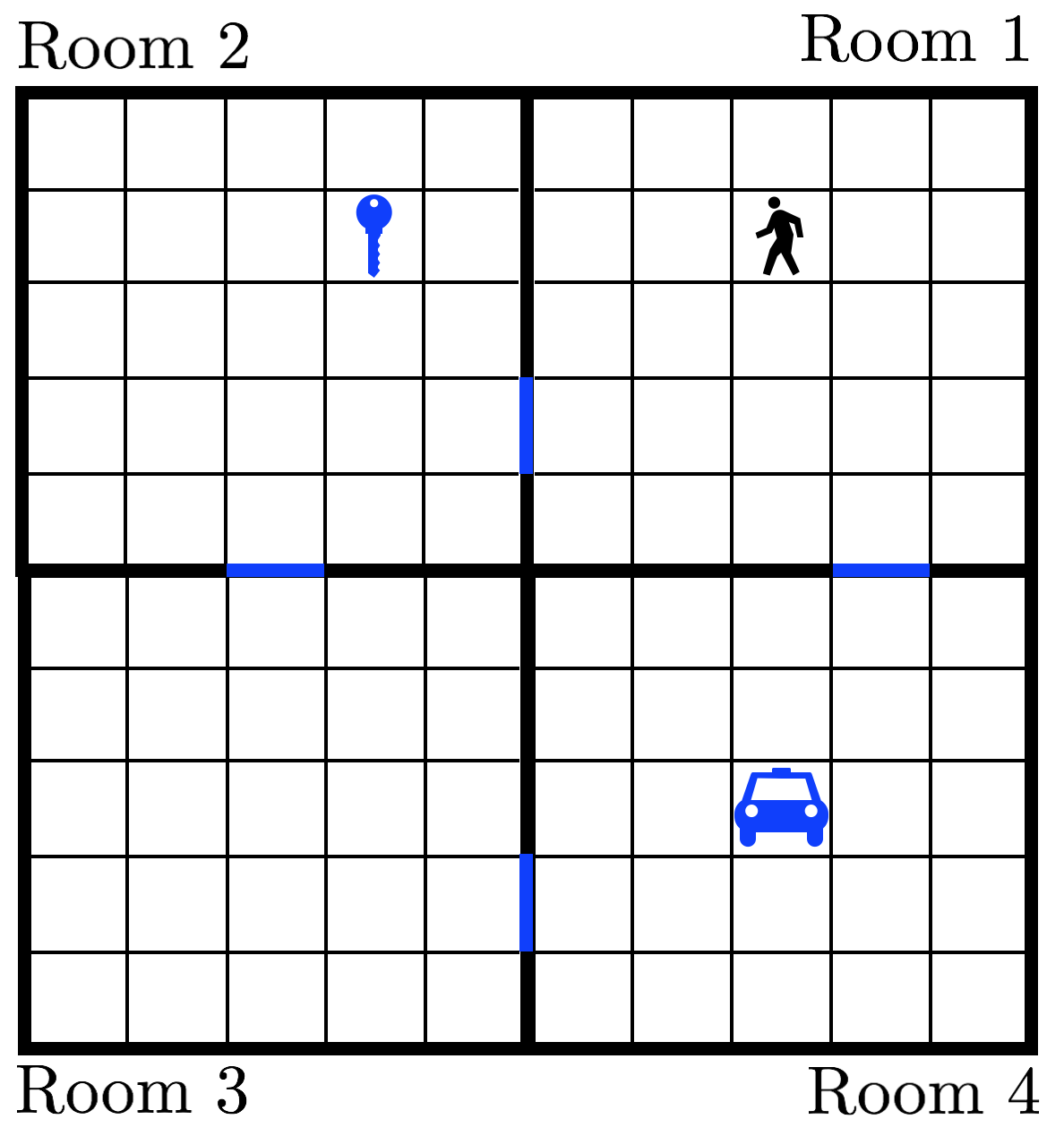} 
	\caption{The rooms task with a key and a car. The agent should explore the \emph{rooms} to first find the key and then find the car. The key and the car can be in any of the 4 rooms in any arbitrary locations. The agent moves either $\mathcal{A} = \{North, South, East, West\}$ on each time step. The agent receives $r=+10$ reward for getting the key and $r=+100$ if it reaches the car with the key. The blue objects on the map --- doorways, key, and car --- are useful \emph{subgoals}.}
	\label{4-room-key-door-car-env}
\end{figure}

The 4-room is a grid-world  environment, consisting of 4 rooms, as shown in Figure \ref{4-room-key-door-car-env}. These rooms are connected through \emph{doorways}. The agent receives the most reward if it navigates in this environment, finds a key, picks up the key, and moves to a car. The agent is initialized in an arbitrary location in an arbitrary room. The location of the key, the car, and the doorways are arbitrary and can vary. This is a variant of the \emph{rooms} task introduced by \citep{Sutton:1999:Option}. The agent receives $r=+10$ reward for reaching the key and $r=+100$ if it moves to the car while carrying the key. The agent can move either $\mathcal{A}=$\{North, South, East, West\} on each time step. Bumping to the wall boundaries is punished with a reward of $r=-2$. There is no reward or punishment for exploring the space. 
Learning in this environment with sparse feedback is challenging for a reinforcement learning agent. To successfully generalize to different environment configurations, the agent should represent knowledge at multiple levels of spatiotemporal abstraction. It should also learn to explore the environment efficiently. The quality of the agent's \emph{policy} depends critically on the location of the doorways, the key, and the car. 

\subsection{Subgoals vs. Options}
\begin{figure*}
	\centering
	\begin{tabular}{cc} 
		\includegraphics[width=0.3\textwidth]{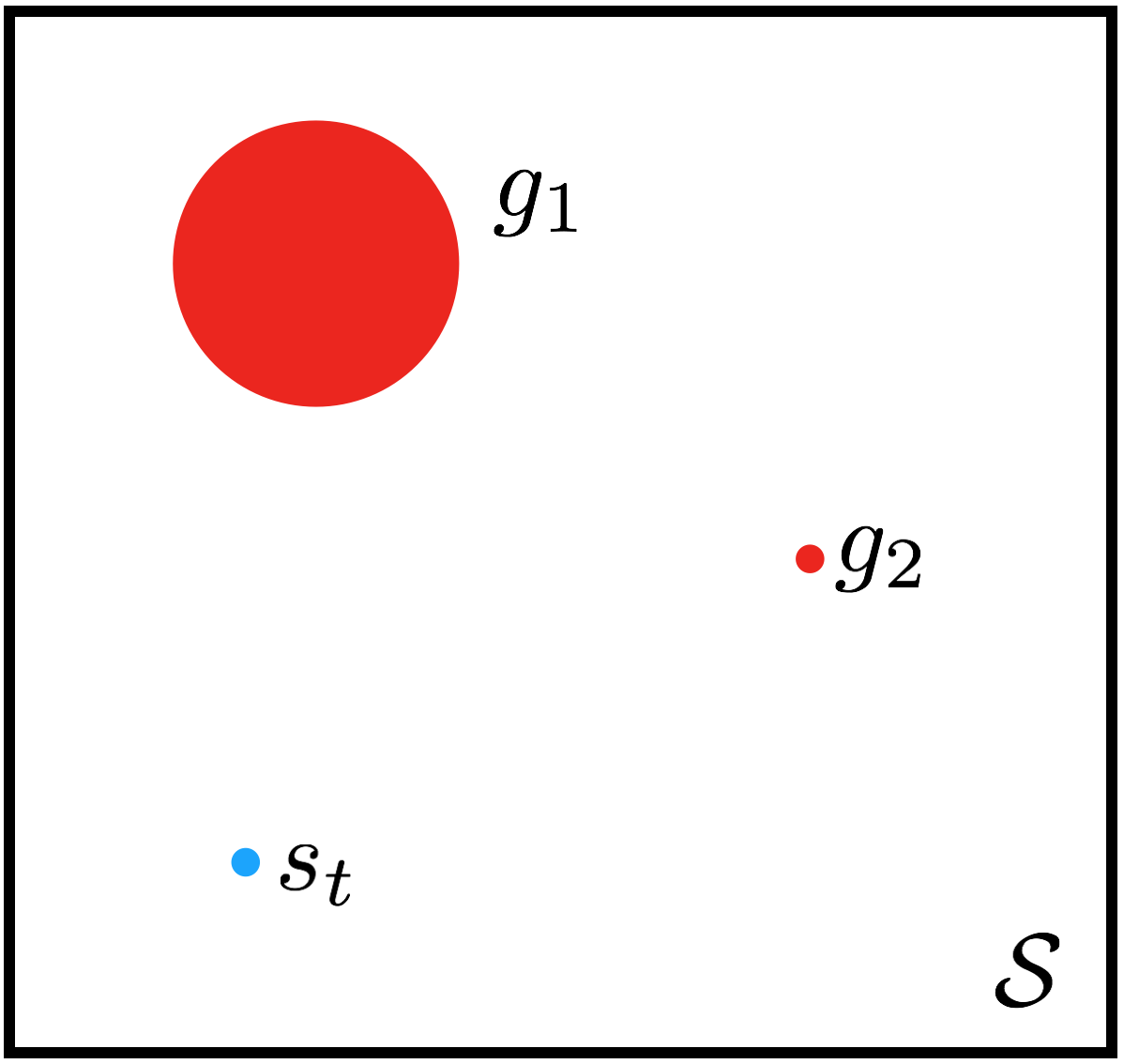} & \includegraphics[width=0.4\textheight]{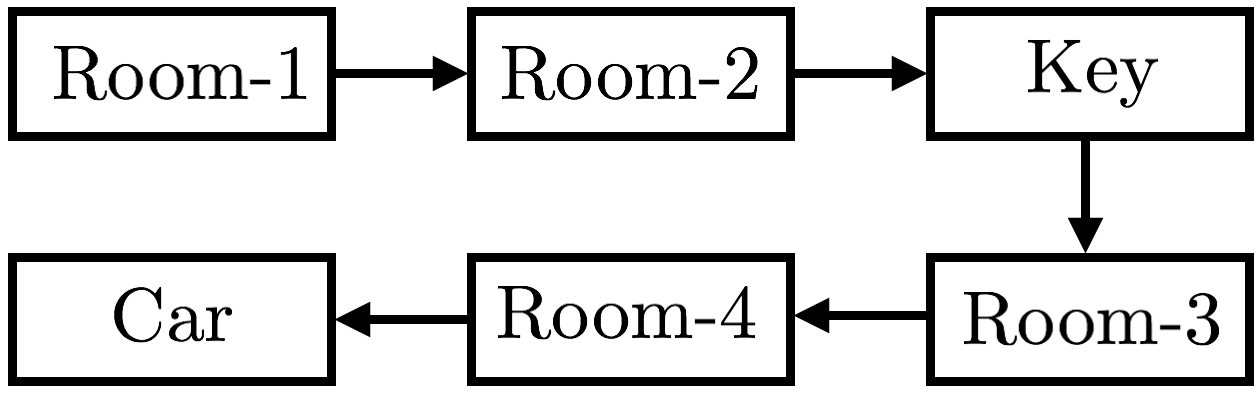}\\
		(a) & (b)
	\end{tabular}
	\caption{(a) The state space $\mathcal{S}$ and the state of the agent $s_t$. The intrinsic goal can be either reaching from $s_t$ to a region or set of states, $g_1 \subset \mathcal{S}$, or to a single state $g_2 \in \mathcal{S}$. (b) An option is a transition from a set of states to another set of states.}
	\label{grf:state-space}
\end{figure*}

Learning to obtain a subgoal is typically easier than learning the full task. Pursuing a \emph{subtask} of ``go to room 2'', which is part of the solution to the full task, is much easier than the 4-rooms task, itself. A \emph{subgoal}, $g$, is a state that must be visited as part of pursuing a major \emph{goal}. The subspace $\mathcal{G} \subseteq \mathcal{S}$ is called the \emph{goal space}, and the members of $\mathcal{G}$ are the candidate subgoals (or goals) that the RL agent might pursue to solve the task. In this task, a good set of subgoals is $\mathcal{G} = $ \verb+{doorways, key, car}+. However, being able to pursue other sets of states can be useful in learning the task. For example, learning how to move from room 2 to room 1 can be a useful skill, and successful execution of this skill is marked by moving to a subset region of state space. (See Figure \ref{grf:state-space}.) 

In some of the HRL literature the term ``option'' is used to describe a temporally abstracted action \citep{Machado:2016:Purposeful,Fox2017MultiLevelDO}. The literature can be confusing, however, as other researchers use this term to describe a subgoal --- a specific state in the state space. In this paper, we use the former notion of option. An option, $o_{ij}$, is a transition from one set of states, $g_i$, to another, $g_j$. For example, going from room 1 to room 2 can be considered an option (i.e., an extended action).

\subsection{Spatiotemporal Hierarchies}
The rooms task has at least two types of hierarchical structure. 
\begin{enumerate}
	\item \emph{Spatial Hierarchy.} The states have similarity structure in terms of belonging to a certain room. This could be captured as a hierarchical relationship between locations and rooms. For example, at the moment captured in Figure \ref{4-room-key-door-car-env}, the key is located in relative location $(x_{key},y_{key})$ in room 2, and the agent is in relative location $(x_{agent},y_{agent})$ in room 1.
	\item \emph{Temporal Hierarchy.} To solve the task, the agent first needs to get the key, and, after accomplishing this subgoal, the agent should move to the car. Thus, the temporal order of subgoals is another dimension of hierarchy.
\end{enumerate}	

\subsection{Hierarchical Reinforcement Learning Subproblems}
\label{sub:hrl-subproblems}
\begin{figure*}
	\centering
	\begin{tabular}{ccc} 
		(a) & (b) & (c)\\
		\includegraphics[width=.3\textwidth]{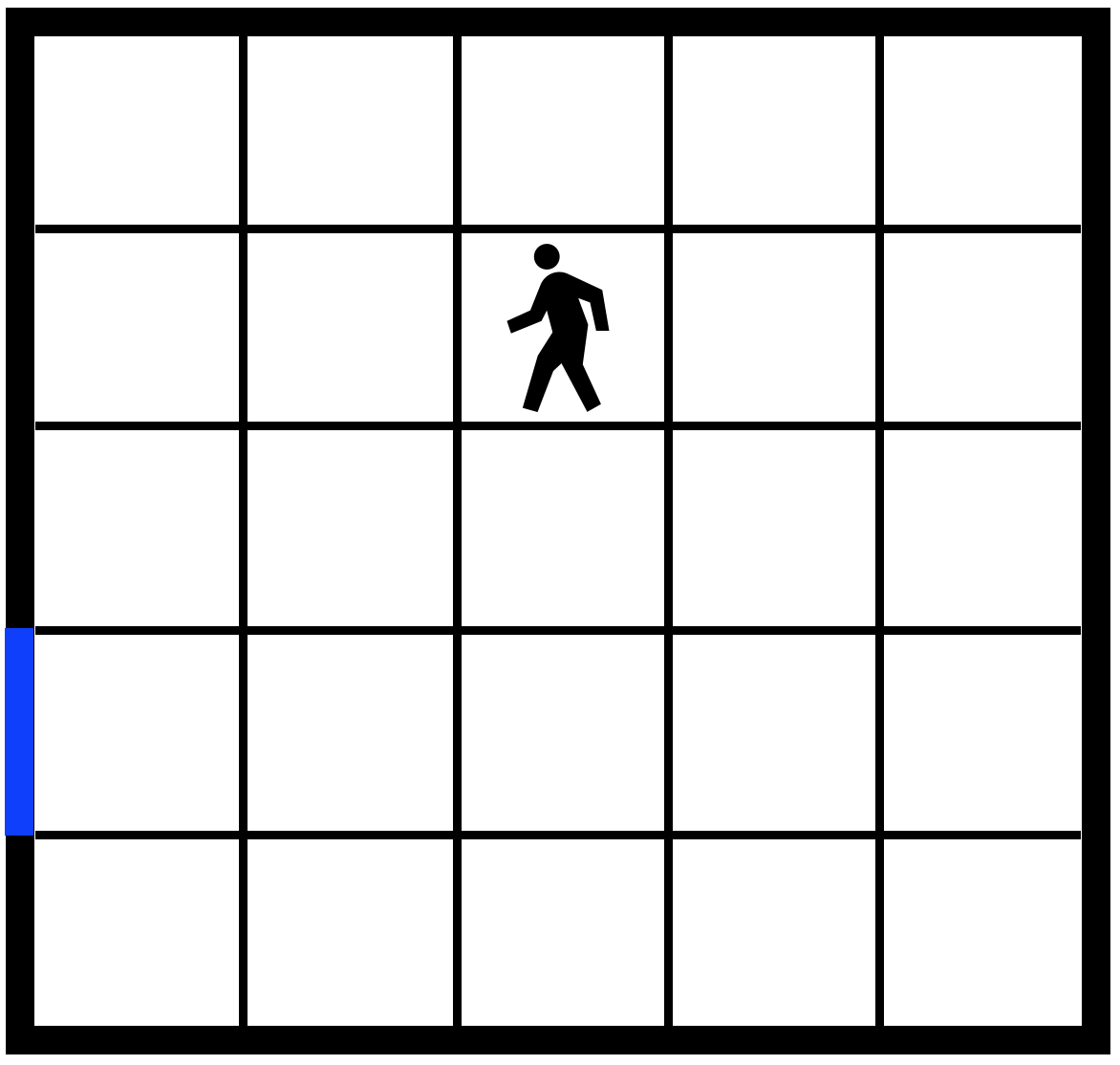} &
		\includegraphics[width=.3\textwidth]{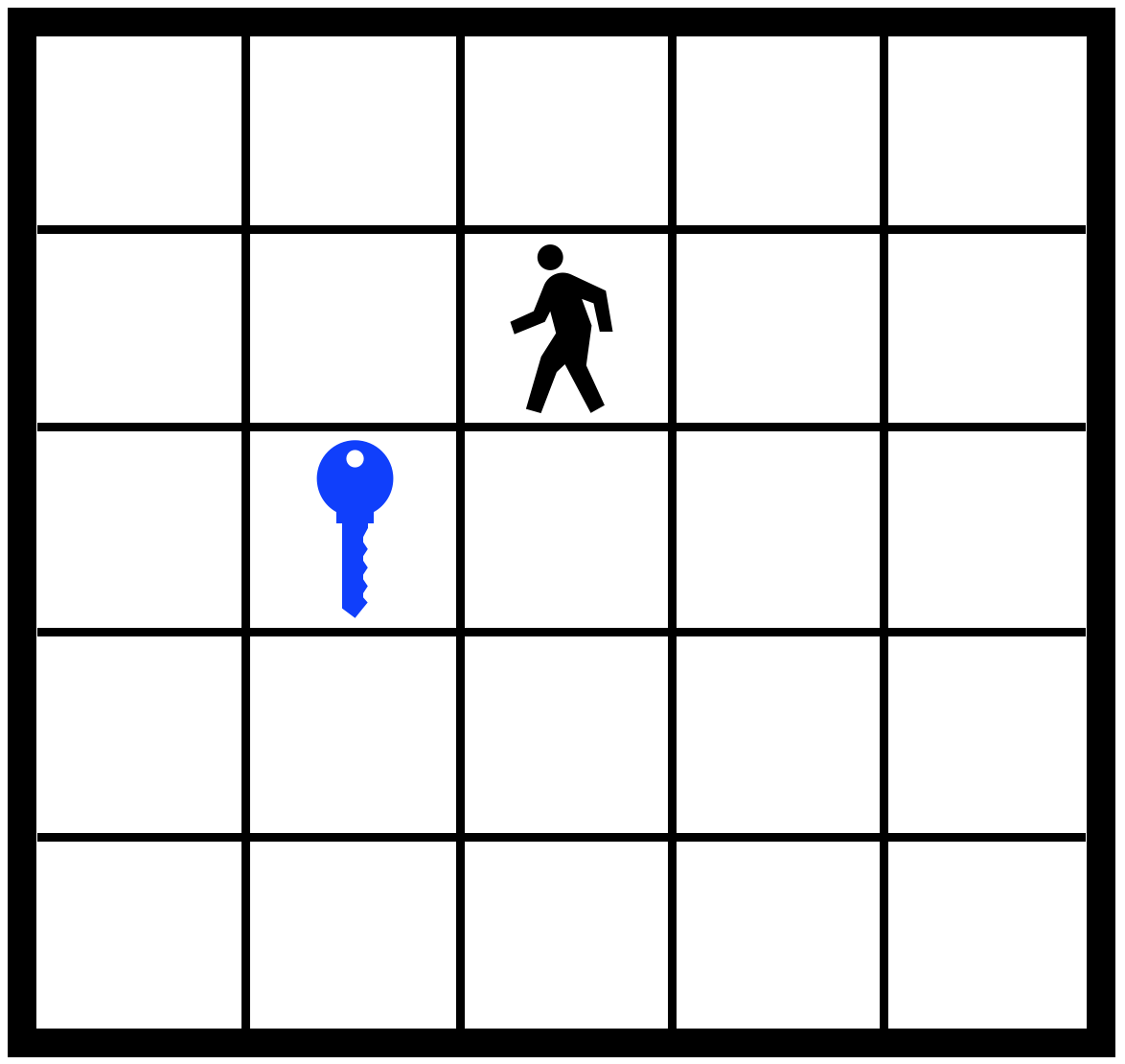} &
		\includegraphics[width=.3\textwidth]{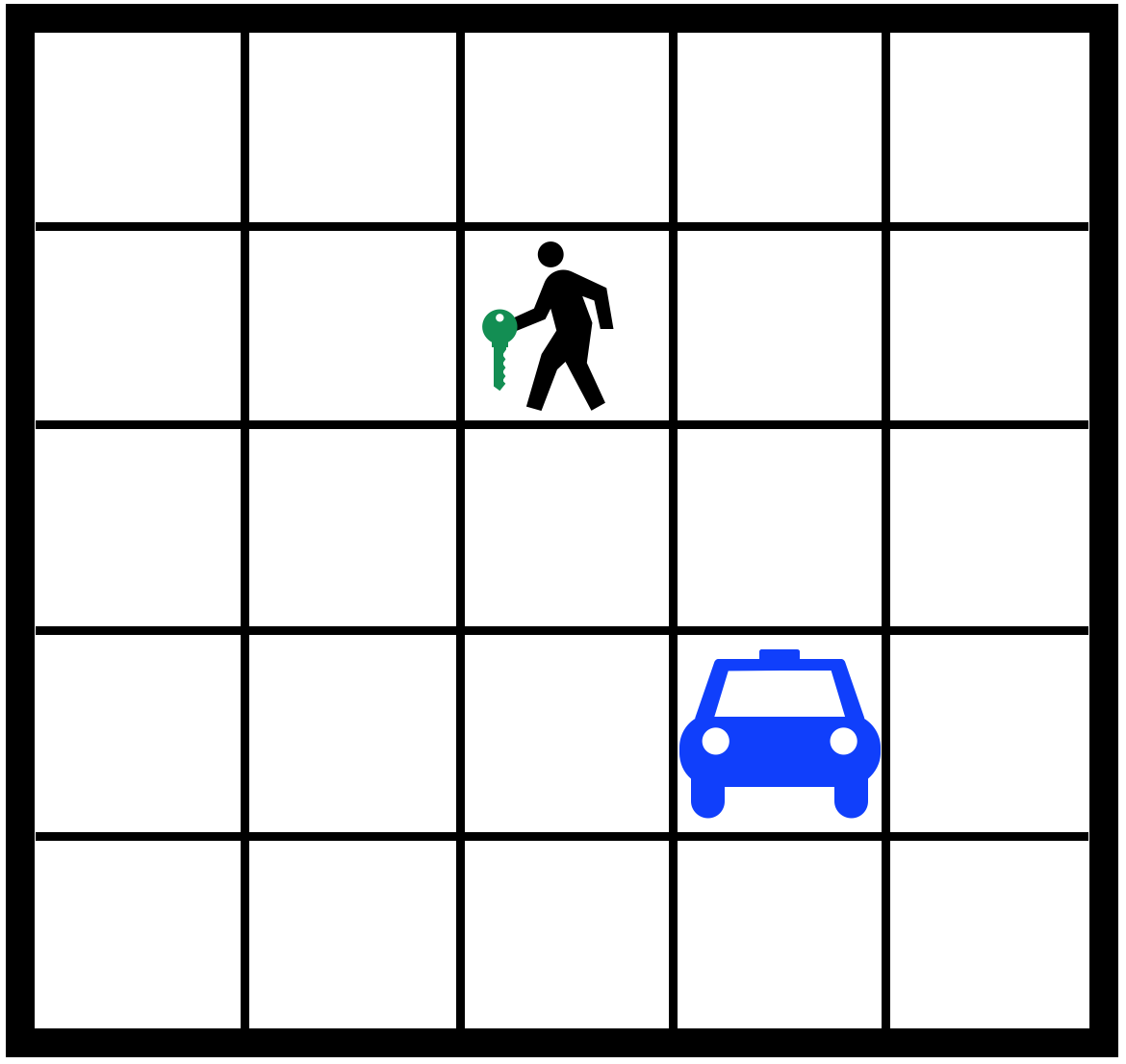} 
	\end{tabular}
	\caption{The rooms task requires the agent to be able to navigate and reach a certain subgoal, $g$, from its current state. (a) Moving to a \emph{doorway}. (b) Moving to the key. (c) Moving to the car. Learning how to space the state space through intrinsic motivation can facilitate learning.}
	\label{grf:room-to-gridworld}
\end{figure*}

The rooms task has both clear skills and clear subgoals. To accomplish each temporal subgoal $g \in $ \verb+{doorways, key, car}+, the agent needs to go to the corresponding location of $g$. (See Figure \ref{grf:room-to-gridworld}.) Learning how to explore the state space to reach any arbitrary location in the given room is a valuable skill. This skill could be reused to reach any of a number of subgoals. This is often called \emph{spacing} and it can be accomplished through \emph{intrinsic motivation}.

One approach to solving the hierarchical reinforcement learning problem is to break down the problem into the following three subproblems:

\paragraph{Subproblem 1: Learning a meta-policy to choose a subgoal.}  
Learning to operate at different levels of abstraction is essential in hierarchical reinforcement learning. In this subproblem, the purpose is training a top-level controller to learn an optimal meta-policy to choose a proper subgoal, $g_t$, form set of candidate subgoals, $\mathcal{G}$, and deliver it to the lower-level controller. We refer to the top-level learner as the \emph{meta-controller}, and the lower-level learner is just called the \emph{controller} \citep{Kulkarni:2016:Meta-Controller}. Formally, the objective is to find a mapping $\Pi: \mathcal{S} \to \mathcal{G}$, which ideally is an optimal meta-policy that maximizes the return. For example, in the rooms task (Figure \ref{4-room-key-door-car-env}), it is expected that the meta-controller chooses a proper subgoal (i.e. room 2) for the agent located in room 1. 

\paragraph{Subproblem 2: Exploring the state space while learning subtaks through intrinsic motivation.} 
Intrinsic motivation refers to learning behavior that is driven by internal rewards. A reinforcement learning agent can be intrinsically motivated to explore its environment and learn about the effects of its actions. The skills learned during this period of exploration are then reused to great effect later to solve many unfamiliar problems very quickly. The agent is
assigned to solve a task of reaching to a subgoal, $g_t$. Formally, the agent should learn an optimal policy $\pi(s_t, g_t)$ for all possible (available) states, $s_t \in \mathcal{S}$ and for all subgoals, $g_t \in \mathcal{G}$. In particular, we present algorithms for intrinsically motivated hierarchical exploration for temporal difference learning. An example of intrinsic motivation is given in Figure \ref{grf:room-to-gridworld}. The meta-controller (top-level
learner) assigns a goal, $g_t$, to the controller (lower-level learner), and the controller should learn how to reach to different locations in the room. Consequently, the agent learns how to navigate in the state space to reach an arbitrary goal.

\paragraph{Subproblem 3: Subgoal discovery.} 
In order to solve subproblems 1 and 2, a proper set of subgoals, $\mathcal{G}$, should be available. This require solving the subgoal discovery problem which is one of major open problems in the hierarchical reinforcement learning literature. Formally, we are interested
in studying methods of learning to discover the proper candidate subgoals, $\mathcal{G}$, from the agent's past experiences memory $\mathcal{D}$. For example, a proper set of subgoals for the rooms task (see Figure \ref{4-room-key-door-car-env}) include the location of doorways, the key and the car. When learning begins, the subgoals are arbitrary, but once they are assigned to the controller, more experiences can be gathered through the process of intrinsic motivation learning. We introduce an unsupervised learning method that can discover the underlying structure in the experience space and use the learned representation to discover a good set of subgoals. Automatic subgoal discovery in model-free hierarchical reinforcement learning is an open problem that is addressed in the proposed HRL framework. 

\section{Meta-controller/Controller Framework}
A straightforward computational approach for temporal abstraction is proposed by \cite{Kulkarni:2016:Meta-Controller} in the meta-controller/controller framework. The agent in this framework makes decisions at two levels of abstraction:
\begin{itemize}
	\item[(a)] The top level module (\emph{meta-controller}) takes the state, $s_t$, as input and picks a new subgoal, $g_t$.
	\item[(b)] The lower level module (\emph{controller}) uses both the state (or a meta-state $\tilde{s}_t$) and the chosen subgoal to select actions, continuing to do so until either the subgoal is reached or the episode is terminated. 
\end{itemize}
If a subgoal is reached, the meta-controller then chooses another subgoal, and the above steps (a-b) repeat. In this paper, we focus on only two levels of hierarchy, but the proposed methods can be expanded to greater hierarchical depth without loss of generality.   

\begin{figure}[hbt!] 
	\centering
	\includegraphics[width=0.6\textwidth]{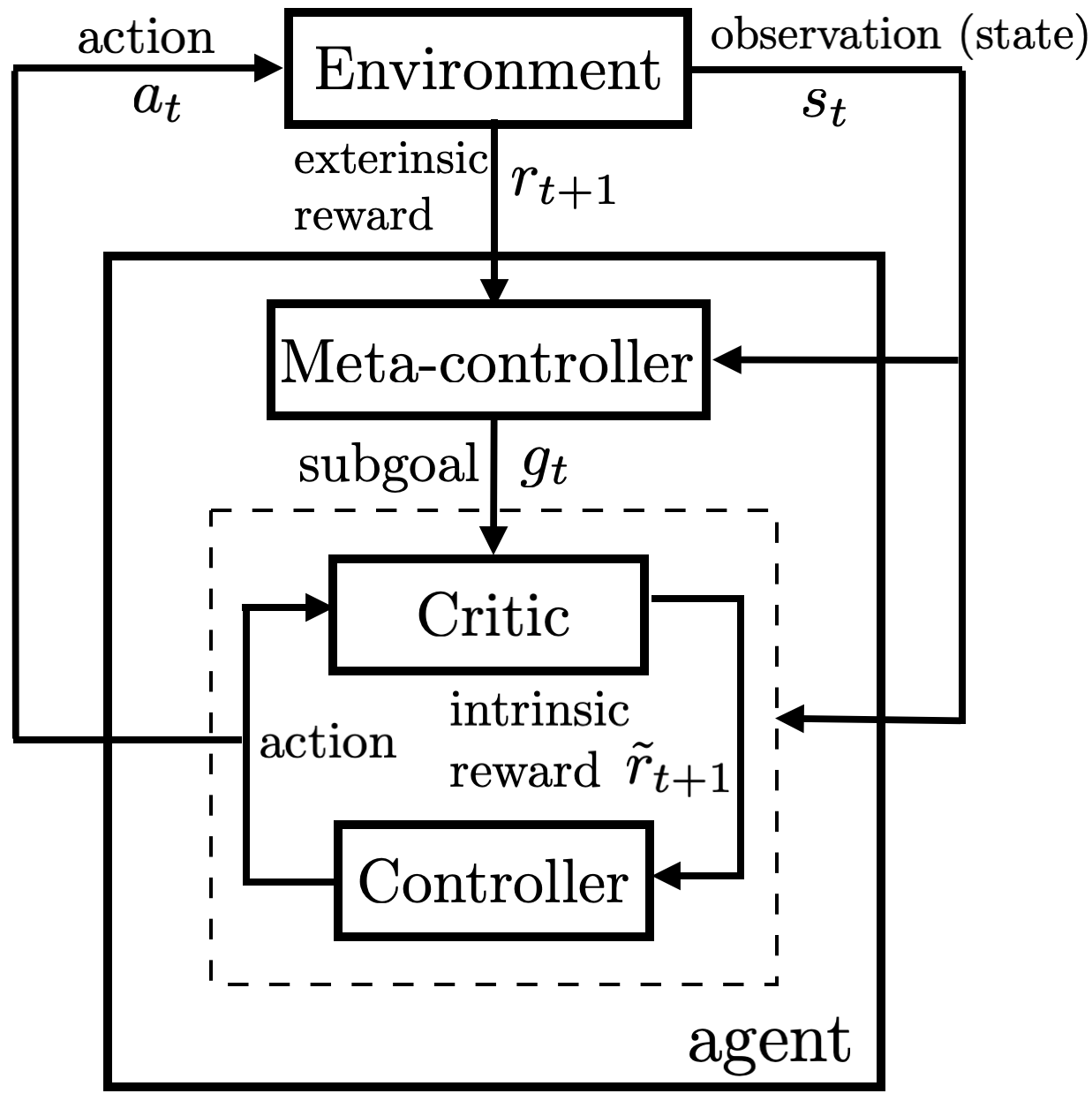}\\
	\caption{The Meta-Controller/Controller framework for temporal abstraction. The agent produces actions and receives sensory observations. Separate networks are used inside the meta-controller and controller. The meta-controller looks at the raw states and produces a policy over goals by estimating the value function $Q(s_t,g_t)$ (by maximizing expected future extrinsic reward). The controller takes states as input, along with the current goal, $(s_t,g_t)$, and produces a policy over actions by estimating the value function $q(s_t, g_t, a_t)$ to accomplish the goal $g_t$ (by maximizing expected future intrinsic reward). The internal critic checks if a goal is reached and provides an appropriate intrinsic reward to the controller. The controller terminates either when the episode ends or when $g_t$ is accomplished. The meta-controller then chooses a new subgoal, and the process repeats. This architecture is adapted from \citep{Kulkarni:2016:Meta-Controller}.}
	\label{grf:meta-controller-controller}
\end{figure}
As shown in Figure \ref{grf:meta-controller-controller}, the agent uses a two-level hierarchy consisting of a controller and a meta-controller. At time step $t$, the meta-controller receives a state observation,
$s=s_t$, from the environment. It has a policy for selecting
a \emph{subgoal}, $g=g_t$, from a set of subgoals, $\mathcal{G}$. The controller then selects actions in an effort to attain the given subgoal.  The objective function for the controller is to maximize cumulative future intrinsic reward 
\begin{align}
\tilde{G_t} = \sum_{t'=t}^{t+T} \gamma^{t'-t} \tilde{r}_t(g),
\end{align}
where $T$ is the maximum length of internal episodes to accomplish the subtask of reaching to subgoal $g$. Similarly, the objective of the meta-controller is to maximize the cumulative extrinsic reward 
\begin{align}
G_t = \sum_{t'=t}^{\mathcal{T}} \gamma^{t'-t} r_t,
\label{eq:retrun-meta}
\end{align}
where $\mathcal{T}$ is a final step. We can use two different $Q$ functions to learn policies for the controller and the meta-controller. The controller estimates the following $Q$ values
\begin{align}
q(s,g,a) = \mathbf{E}_{\pi_{ag}} \big[ \tilde{G}_t | s_t=s,g_t=g,a_t=a \big],
\end{align} 
where $g$ is the given subgoal in state $s$ and $\pi_{ag} = P(a|s,g)$ is the action policy.
Similarly, the meta-controller estimates the following $Q$ values
\begin{align}
Q(s,g) = \mathbf{E}_{\pi_{g}} \big[ G_t | s_t=s,g_t=g\big],
\end{align} 
where $\pi_g$ is the policy over subgoals. For example the optimal meta-policy for the rooms task is depicted in Figure \ref{grf:state-space} (b). It's important to note that the meta-controller experiences transitions, $(s_t,g_t,G_{t:t+T},g_{t+T})$, at a slower time-scale than the controller's transitions, $(s_t, a_t, g_t, \tilde{r}_t, s_{t+1})$. Note that $G_{t:t+T}$ is the return (cumulative external reward) in Equation \ref{eq:retrun-meta} for one episode of the controller with a length $T$.

In our implementation, the policy arises from estimating the value of
each subgoal, $Q(s,g;\mathcal{W})$, and selecting the goal of highest
estimated value. With the current subgoal selected, the controller uses its policy to select an action, $a \in \mathcal{A}$, based on the current state, $s$, and the current subgoal, $g$. In our implementation, this policy involves selecting the action that results in the highest estimate of the controller's value function, $q(s,g,a;w)$. Actions continue to be
selected by the controller while an internal critic monitors the
current state, comparing it to the current subgoal, and delivering an 
appropriate \emph{intrinsic reward}, $\tilde{r}$, to the controller on
each time step. Each transition experience, $(s,g,a,\tilde{r},s')$, is
recorded in the controller's experience memory set, $\mathcal{D}_{1}$,
to support learning. When the subgoal is attained, or a maximum amount
of time has passed, the meta-controller observes the resulting state,
$s_{t'}=s_{t+T+1}$, and selects another subgoal, $g'=g_{t+T+1}$, at
which time the process repeats, but not before recording a transition
experience for the meta-controller, $(s,g,G,s_{t'})$ in the
meta-controller's experience memory set, $\mathcal{D}_{2}$. The
parameters of the value function approximators are adjusted based on
the collections of recent experiences. 

For training the meta-controller value function, we minimize a loss function based on the reward received from the environment:
\begin{align}
\mathcal{L}(\mathcal{W}) \triangleq \mathbb{E}_{(s,g,G,s_{t'})\sim \mathcal{D}_{2}}\big[ \big(G + \gamma \max_{g'} Q(s_{t'},g';\mathcal{W}) - Q(s,g;\mathcal{W}) \big)^2 \big],
\label{eq:meta-loss}
\end{align}
where $G=\sum_{t'=t}^{t+T} \gamma^{t'-t} r_{t'}$ is the accumulated
external reward (return) between the selection of consecutive
subgoals. The term, $\mathcal{Y} = G+\gamma \max_{g'} Q(s_{t'},g';\mathcal{W})$, in \eqref{eq:meta-loss} is the target value for the expected return at the time that the meta-controller selected subgoal $g$.
The controller improves its subpolicy, $\pi(a|s,g)$, by learning its value function, $q(s,g,a;w)$,
over the set of recorded transition experiences. The controller
updates its value function approximator parameters, $w$, so as to
minimize its loss function:
\begin{align}
L(w) \triangleq \mathbb{E}_{(s,g,a,\tilde{r},s')\sim \mathcal{D}_{1}}\big[ \big(\tilde{r} + \gamma \max_{a'} q(s',g,a';w) - q(s,g,a;w) \big)^2 \big].
\label{eq:loss-intrinsic}
\end{align}

The hierarchical reinforcement learning algorithm for meta-controller and controller learning is given in Algorithm \ref{Algo:meta-controller-controller}.
\begin{algorithm}[hbt!] 			
	\begin{algorithmic}
		\State Specify Subgoals space $\mathcal{G}$
		\State Initialize $w$ in $q(s,g,a;w)$ 
		\State Initialize $\mathcal{W}$ in $Q(s,g;\mathcal{W})$.
		\State Initialize experience memories $\mathcal{D}_1$ and $\mathcal{D}_2$
		\For{ episode $=1,\dots,M$} 
		\State Initialize state $s_0 \in \mathcal{S}$, $s\gets s_0$
		\State $G \gets 0$
		\State $g \gets$\texttt{EPSILON-GREEDY}$(Q(s,\mathcal{G};\mathcal{W}),\epsilon_2)$
		\Repeat{ for each step $t = 1,\dots,T$} 					
		\State compute $q(s,g,a;w)$
		\State $a\gets$\texttt{EPSILON-GREEDY}$(q(s,g,\mathcal{A};w),\epsilon_1)$
		\State Take action $a$, observe $s'$ and external reward $r$ 
		\State Compute intrinsic reward $\tilde{r}$ from internal critic
		\State Store controller's intrinsic experience, $(s,g,a,\tilde{r},s')$ to $\mathcal{D}_1$	
		\State Sample $J_1 \subset \mathcal{D}_1$ and compute $\nabla L$
		\State Update controller's parameters, $\quad w \gets w - \alpha_1 \nabla L$
		\State Sample $J_2 \subset \mathcal{D}_2$ and compute $\nabla \mathcal{L}$
		\State  Update meta-controller's parameters, $\quad \mathcal{W} \gets \mathcal{W} - \alpha_2 \nabla \mathcal{L}$ 
		\State $s \gets s',\quad G \gets G + r$
		\State Decay exploration rate of controller $\epsilon_1$
		\Until{$s$ is terminal or subgoal $g$ is attained}			
		\State Decay exploration rate of meta-controller $\epsilon_2$
		\State Store meta-controller's experience, $(s_0,g,G,s')$ to $\mathcal{D}_2$			
		\EndFor		
	\end{algorithmic}
	\caption{Meta-Controller and Controller Learning}
	\label{Algo:meta-controller-controller}
\end{algorithm}

\section{Intrinsic Motivation Learning}
Intrinsic motivation learning is the core idea behind the learning of
value functions in the meta-controller and the controller. In some
tasks with sparse delayed feedback, a standard RL agent cannot
effectively explore the state space so as to have a sufficient number
of rewarding experiences to learn how to maximize rewards. In
contrast, the intrinsic critic in our HRL framework can send much more
regular feedback to the controller, since it is based on attaining
subgoals, rather than ultimate goals. As an example, our
implementation typically awards an intrinsic reward of $+1$ when the
agent attains the current subgoal, $g$, and $-1$ for any other state
transition. Successfully solving a difficult task not only depends on such an intrinsic motivation learning mechanism, but also on the
meta-controller's ability to learn how to choose the right subgoal for
any given state, $s$, from a set of candidate
subgoals. Indeed, identifying a good set of candidate subgoals is an
additional prerequisite for success.

\begin{figure}
	\centering
	\includegraphics[width=0.5\textwidth]{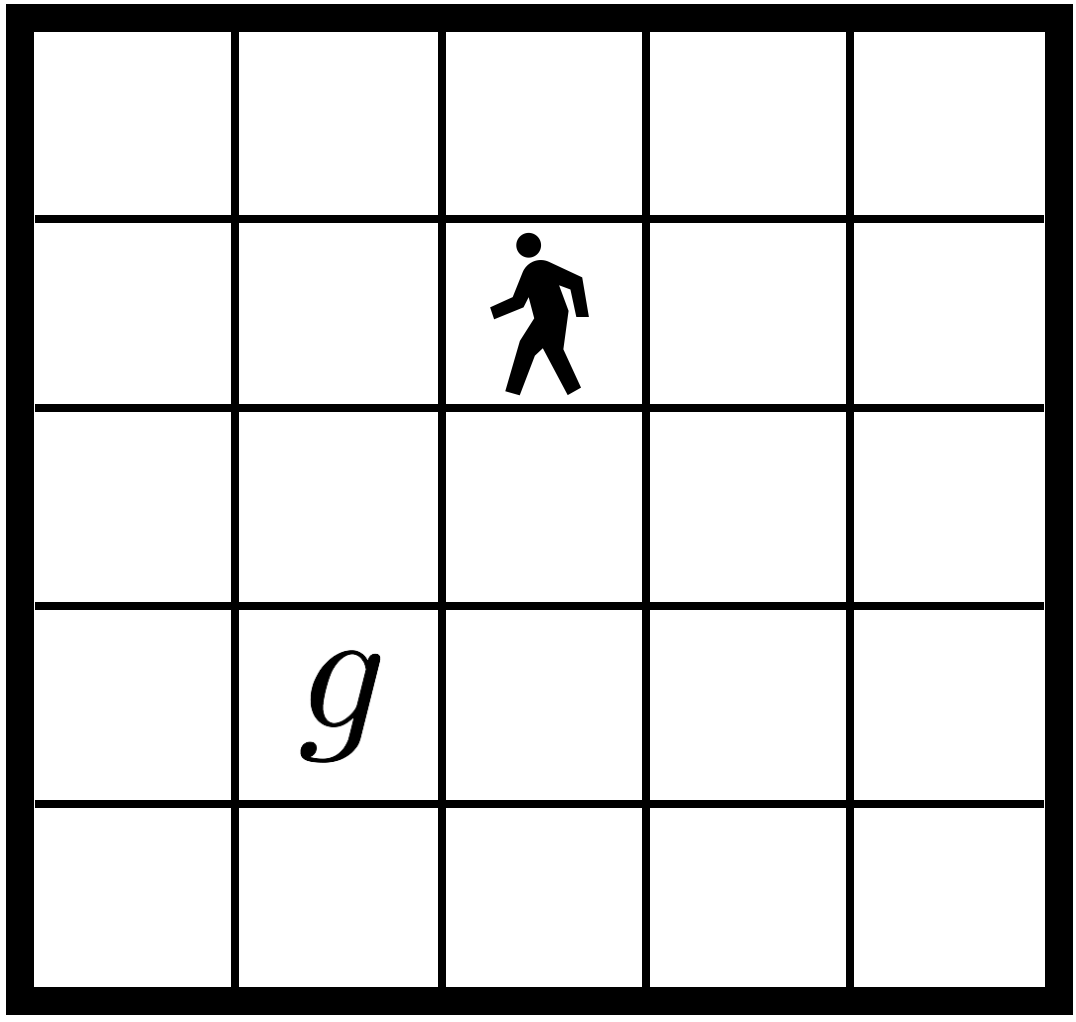}
	\caption{Grid-world task with a dynamic goal. At beginning of each episode an oracle chooses an arbitrary goal, $g \in \mathcal{S}$. The agent is initialized in a random location. On each time step, the agent has four action choices, $\mathcal{A}=$ \{North, South, East, West\}. The agent receives $\tilde{r}=+1$ reward for successful episodes, reaching the goal, $g$. Bumping into the wall produces a reward of $\tilde{r}=-2$. There is no external reward or punishment from the environment for exploring the space.}
	\label{grf:grid-world-goal}
\end{figure}

Developing skills through intrinsic motivation has at least two benefits: (1) exploration of large scale state spaces, and (2) enabling the reuse of skills in varied environments. Navigation in the rooms task requires the agent to learn the skills to reach the doorways, key, and car (see Figure \ref{grf:room-to-gridworld}). These skills are acquired by learning to achieve subgoals that are provided by the meta-controller. The spacing of the state space can be done as a pretraining step or simultaneously with meta-policy training. In any case, a goal should be provided to the controller. This goal can be a random state or a region of state space (see Figure \ref{grf:state-space})). For now, we assume that the subgoal, $g \in \mathcal{G}$, is provided by an oracle (standing in for the meta-controller), and we focus only on learning to achieve this subgoal. The controller receives the state, $s_t$, and subgoal, $g_t$, as inputs and takes an action, $a_t$. This results in the sensing of the next state and the receipt of an intrinsic reward signal, $\tilde{r}_{t+1}$, from the internal critic. (See Figure \ref{grf:meta-controller-controller}.) The subgoal, $g_t$, remains the same for some time, $T$. There have been some studies concerning the appropriate structure of the intrinsic reward. We use the following form
\begin{align}
\tilde{r}_{t+1} = \begin{cases}
\min(r_{t+1},-1) &\textrm{if } s_{t+1} \textrm{ is not terminal } \\
+1 &\textrm{if } s_{t+1} \textrm{ achieves the goal, } g_t
\end{cases}
\label{eq:general-intrinsic-reward}
\end{align} 
Other intrinsic reward functions might be considered. Indeed, the nature and origin of good intrinsic reward functions is an open question in reinforcement learning. As an attempt to solve Subproblem 1, we try to solve the task of navigation in a grid-world given a subgoal location, $g$. (See Figure \ref{grf:grid-world-goal}.) 
For intrinsic motivation we can define a reward function like that above in Equation \ref{eq:general-intrinsic-reward}. The algorithm for intrinsic motivation learning for a random meta-controller is given in Algorithm \ref{Algo:intrinsic-motivation}.     
\begin{algorithm}
	\begin{algorithmic}
		\State Specify Subgoals space $\mathcal{G}$
		\State Initialize $w$ in $q(s,g,a;w)$ 
		\State Initialize controller's experience memory, $\mathcal{D}_1$
		\State Initialize agent's experience memory, $\mathcal{D}$
		\For{ episode $=1,\dots,M$} 
		\State Initialize state $s_0 \in \mathcal{S}$, $s\gets s_0$
		\State Select a random subgoal $g$ from $\mathcal{G}$
		\Repeat{ for each step $t = 1,\dots,T$} 					
		\State compute $q(s,g,a;w)$
		\State $a\gets$\texttt{EPSILON-GREEDY}$(q(s,g,\mathcal{A};w),\epsilon_1)$
		\State Take action $a$, observe $s'$ and external reward $r$ 
		\State Compute intrinsic reward $\tilde{r}$ from internal critic
		\State Store controller's intrinsic experience, $(s,g,a,\tilde{r},s')$ to $\mathcal{D}_1$
		\State Store agent's experience, $(s,a,s',r)$ to $\mathcal{D}$	
		\State Sample $J_1 \subset \mathcal{D}_1$ and compute $\nabla L$
		\State Update controller's parameters, $\quad w \gets w - \alpha_1 \nabla L$
		\State $s \gets s'$
		\State Decay exploration rate of controller $\epsilon_1$
		\Until{$s$ is terminal or subgoal $g$ is attained}						
		\EndFor		
	\end{algorithmic}
	\caption{Intrinsic Motivation Learning}
	\label{Algo:intrinsic-motivation}
\end{algorithm}

\section{Experiment on Intrinsic Motivation Learning}
Here, we want to show why intrinsic motivation can be useful through transferring knowledge and reusing skills. It is important to note that, we trained an agent in a single room (a grid-world environment) to learn the navigation task (see Figure \ref{grf:grid-world-goal}). In the original rooms task in Figure \ref{4-room-key-door-car-env}, there are walls between rooms and specific doorways between rooms. We assume that there is an oracle in the meta-controller that provides the subgoals and a transformation from the real state to the input state to the state-goal network in Figure \ref{grf:state-goal-kwta}. We make this assumption here for the purpose of simplicity in the implementation and in order to prove the advantages of the intrinsic motivation in hierarchical reinforcement learning. We will revisit the intrinsic motivation problem once again in Section \ref{sec:unified-hrl} and solve the rooms task using a Unified Model-Free HRL framework, without having access to the mentioned oracle. For now, let's assume that there is an oracle (instead of a meta-controller) that can transform the real external state of the agent in the rooms task into the location of the agent in the current room, providing this transformed location as the state-goal network input. This makes it possible to reuse the learned navigation skill to solve the rooms task (which is consists of four rooms). 
\subsection{Training the State-Goal Value Function}
\begin{figure}[hbt!] 
	\centering
	\includegraphics[width=0.7\textwidth]{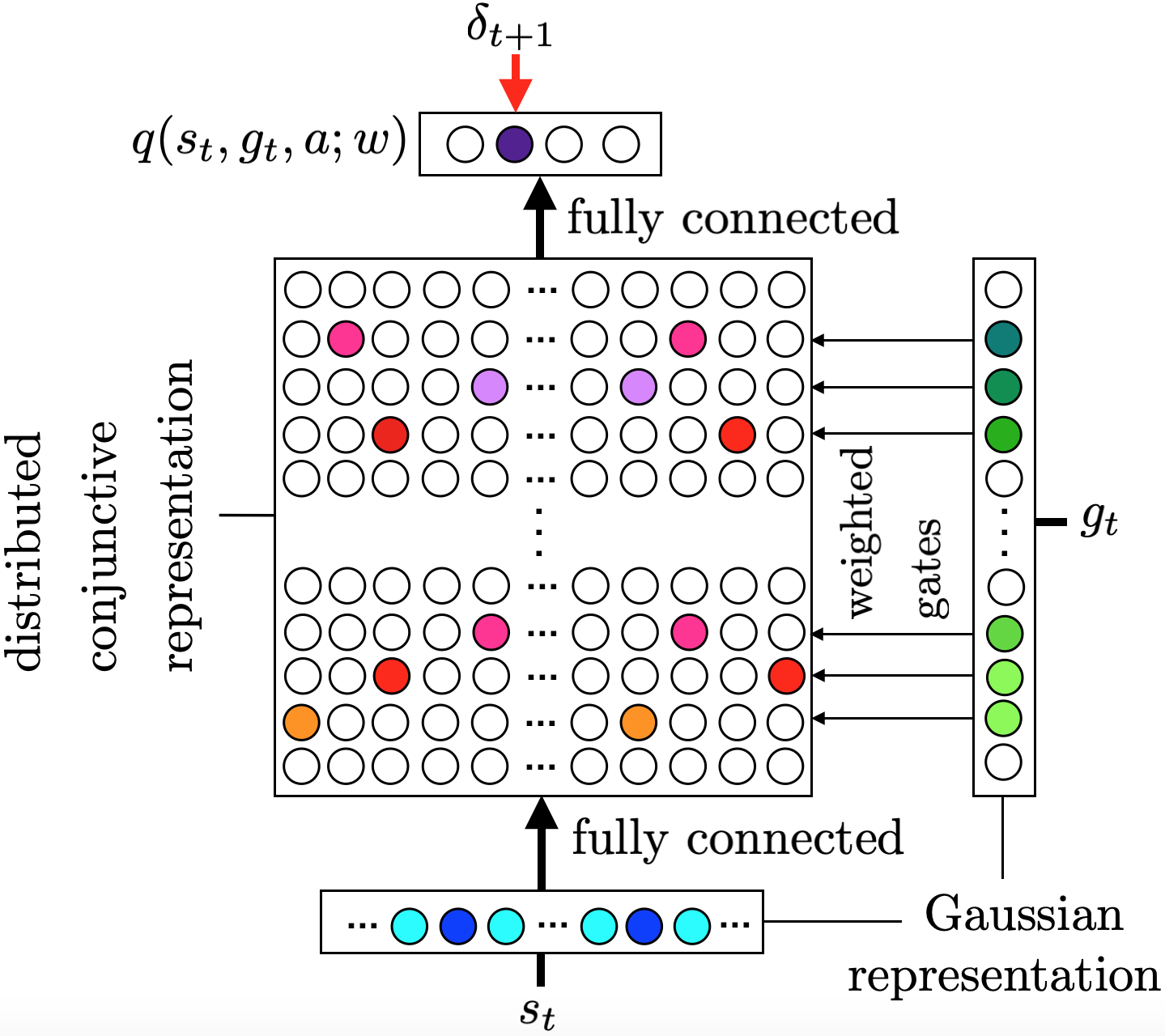}
	\caption{The state-goal neural network architecture used to approximate the value function for the controller, $q(s,g,a;w)$. The function takes the state, $s_t$, and the goal, $g_t$, as inputs. The first layer produces the Gaussian representation separately for $s_t$ and $g_t$. The state representation is connected fully to the hidden layer, and the $k$-Winners-Take-All mechanism produces a sparse representation for $s_t$. The goal representation is connected only to the corresponding row of units. We assume that an oracle in the meta-controller transform the state in rooms task to a proper state for the state-goal netwrok that is trained on the navigation in the  gridworld (single room) task. }
	\label{grf:state-goal-kwta}
\end{figure}

Here, we introduce the neural network architecture that we use to approximate $q(s,g,a;w)$ while training the controller depicted in Figure \ref{grf:state-goal-kwta}.  

The controller $Q$-function, $q(s,g,a;w)$, takes the state, $s_t$, and the goal, $g_t$, as inputs. The first layer produces the Gaussian representation separately for $s_t$ and $g_t$. Let's denote the Gaussian representation of $s_t$ by $\hat{s}_t$ and the goal one by $\hat{g}_t$. The Gaussian representation for $s_t$ similar to the one discussed in \cite{Rafati-Noelle:2015:CSC,Rafati-Noelle:2017:CCCN}, $\hat{s}_t$, is connected fully to the hidden layer, with the connection weight matrix being $w^{(1)}$. The $k$-Winners-Take-All mechanism ($10\%$ of hidden units) produces a sparse representation for the state, $s_t$. The subgoal input, $\hat{g}_t$, is connected to the hidden layer with a gate layer. This mechanism was included in hopes of avoiding catastrophic interference during reinforcement learning. The hidden layer is connected fully to the output units, with the weight matrix being $w^{(2)}$. The network is trained in a manner similar to standard backpropagation, with a forward pass determining activations and a backward pass performing error credit assignment. This training process is summarized in Algorithm \ref{Algo:state-goal-feed-forward}. The grid-world room is discretized into $5 \times 5$ windows. The number of the hidden units in each row of $w^{(1)}$ is 50 and the total number of hidden units are 1250 (considering that there are 25 columns for each row). We assume that the oracle standing in for the meta-controller handles the  transformation of the state in the rooms task to the input for the state-goal network.   
\begin{algorithm}
	\textbf{Forward pass.} Computing $q(s,g,a;w)$
	\begin{algorithmic}
		\State initialize $q^{output} \gets$\texttt{zeros-like}$(a)$
		\State compute $\hat{s}$, Gaussian representation of $s$
		\State compute $\hat{g}$ Gaussian representation of subgoal $g$  
		\State find $id_g$, effective gates indices for which $\hat{g} > 0.1$ 
		\State compute net input matrix for $id_g$, $net = w^{(1)}_{id_g} \hat{s}$
		\ForAll{$i$ in $id_g$}
		\State  compute $net^{i}_{kWTA} \gets$\texttt{kWTA}$(net^{i},k)$
		\State compute activity $h^{i} \gets$ \texttt{sigmoid}$(net^{i}_{kWTA})$
		\State compute $q^{i}$ = $w^{(2)}_i h^i$
		\State $q^{output} \gets q^{output} + q^{i}$
		\EndFor \\
		\Return $q^{output}$
		
	\end{algorithmic}
	\textbf{Backpropagation. } Updating $w$ given TD error ${\delta}$.
	
	\begin{algorithmic}
		\ForAll{$i$ in $id_g$}
		\State 	compute propagated error for hidden units  ${\delta}^i_j \gets {\delta} w^{(2)}_{i,a} \odot h^i \odot (1 - h^i)$
		\State $w^{(2)}_{i,a} \gets  w^{(2)}_a + \alpha {\delta} h^i $
		\State $w^{(1)}_i \gets w^{(1)}_i + \alpha \hat{s} {\delta}^i_j$
		\EndFor 						
	\end{algorithmic}
	\caption{Forward pass, and backpropagation for network in Figure \ref{grf:state-goal-kwta}}
	\label{Algo:state-goal-feed-forward}
\end{algorithm}

\subsection{Intrinsic Motivation Performance Results}
We tested the performance of our approach in the context ofa dynamic rooms environment, with the agent rewarded for solving the key-car grid world task. We used the four following specific tasks to test the learning performance. 

\begin{figure} 
	\begin{center}
		\begin{tabular}{cc} 
			\includegraphics[width=0.4\textwidth]{figures/grid-world-key.png} & 
			\includegraphics[width=0.4\textwidth]{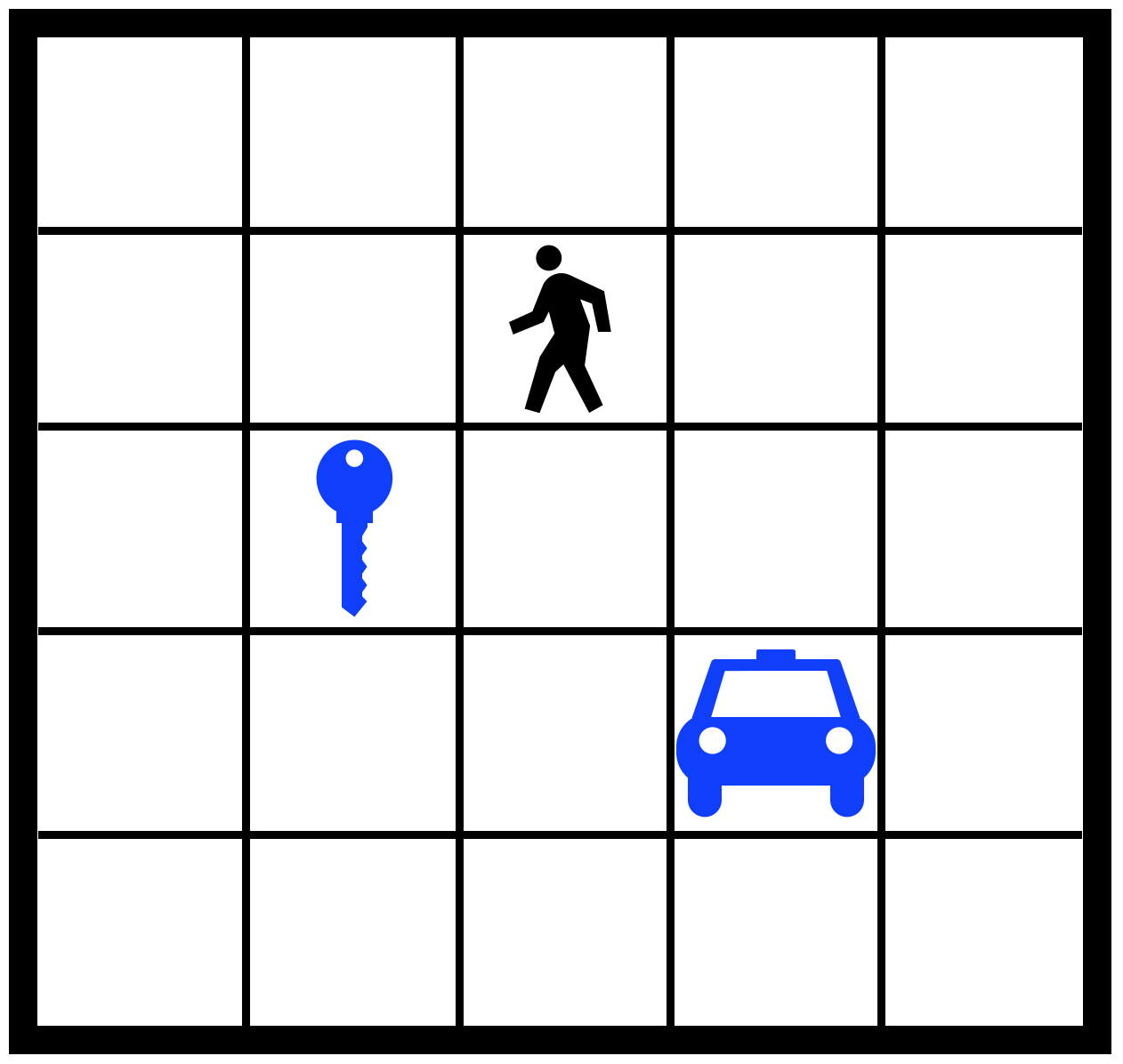}\\
			(a) & (b)\\
		\end{tabular}
	\end{center}
	\caption{In general, the agent received $r=+10$ reward for moving to the key and $r=+100$ if it then moved to the car. On each time step, the agent had four action choices $\mathcal{A}=$ \texttt{\{North, South, East, West\}}. Bumping to the wall produced a reward of $r=-2$. There was no other reward or punishment from the environment for exploring the space. (a) Key task: agent needs to reach to the location of key. (b) Key-Car task: agent should first reach to the key and then to the car. }
	\label{grid-world-key-car-env}
\end{figure}

\begin{figure*} 
	\centering
	\begin{tabular}{cc}
		\includegraphics[width=.43\textwidth]{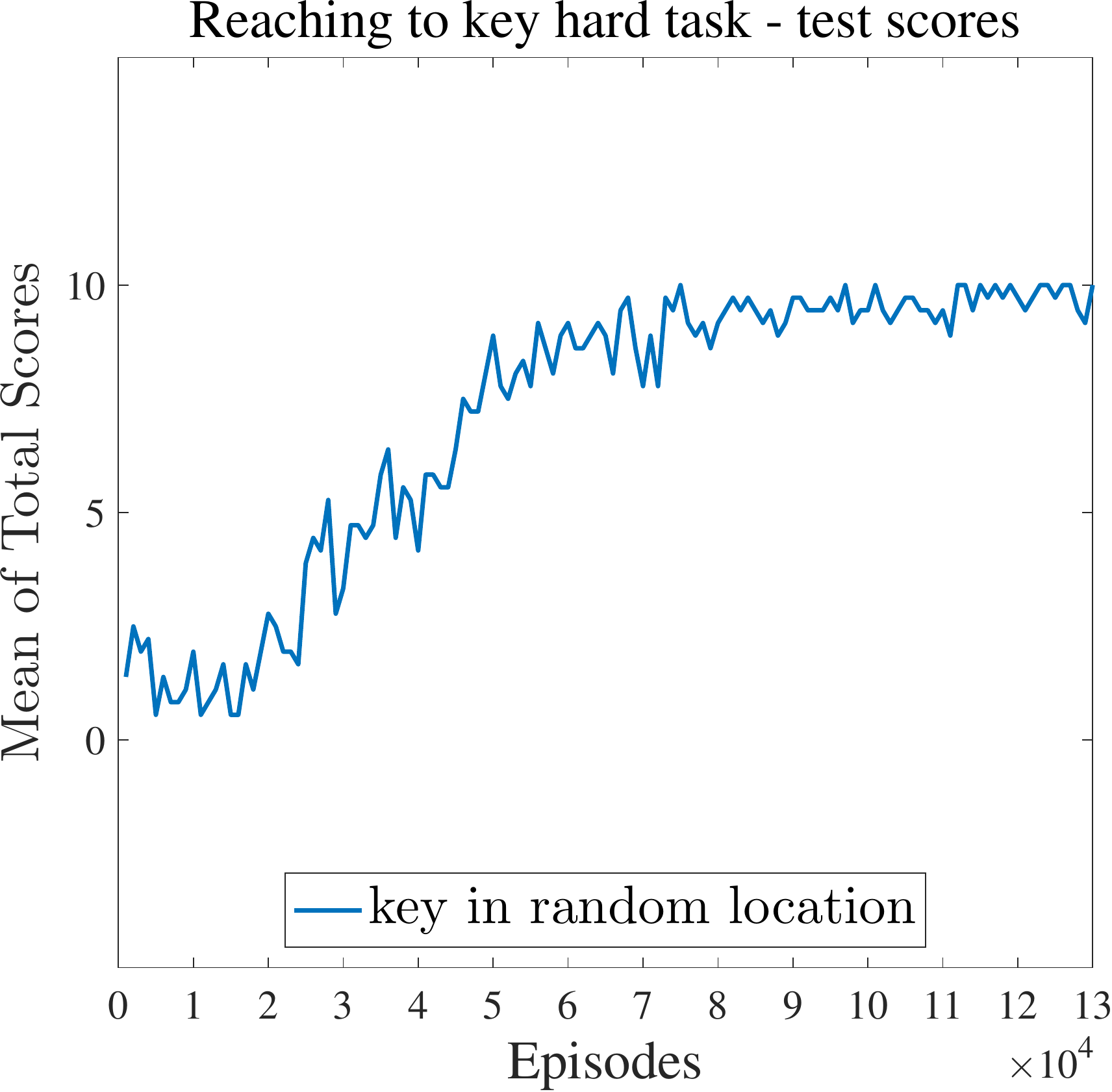} &
		\includegraphics[width=.44\textwidth]{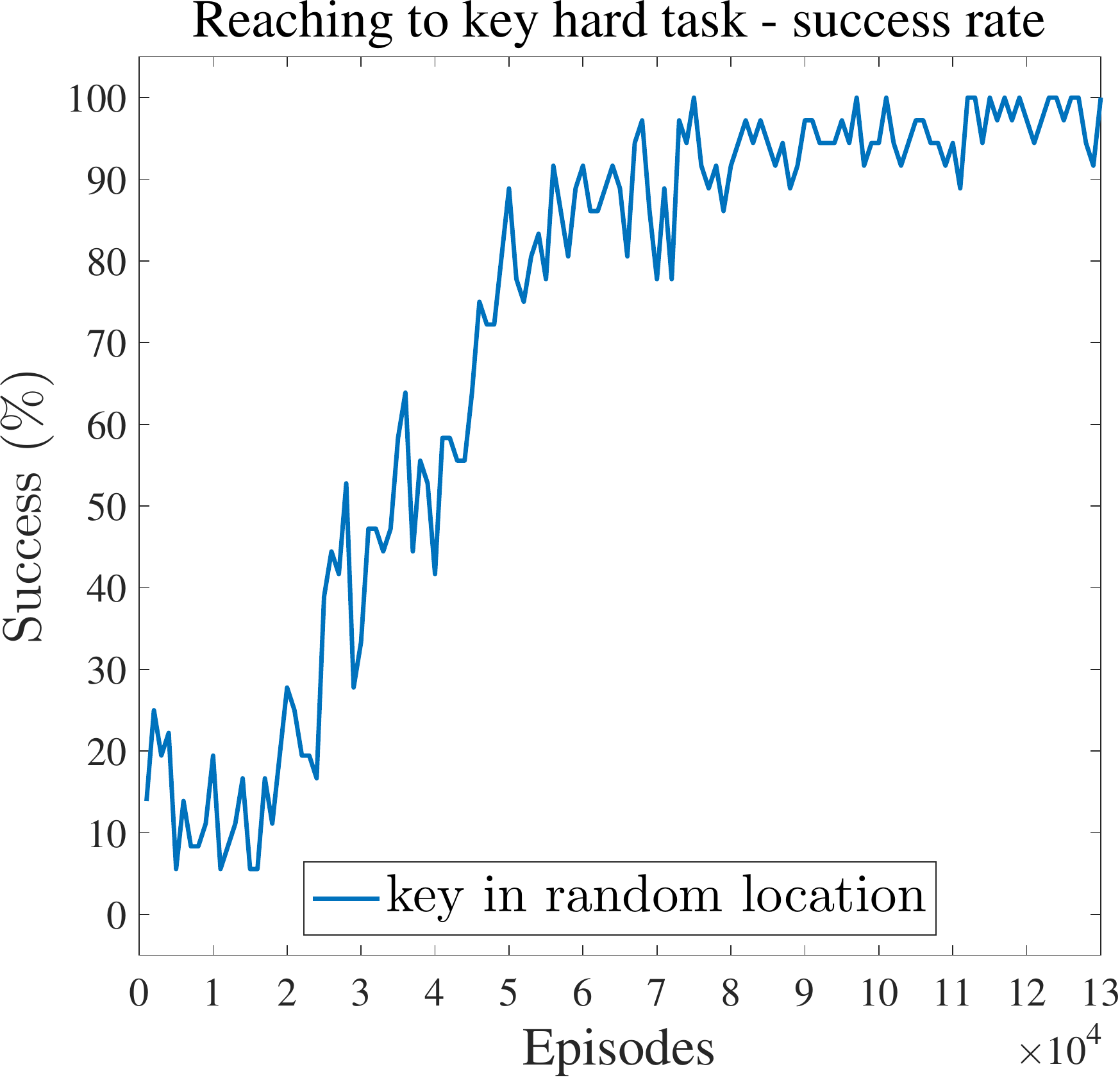}\\
		(a) & (b)\\
		\includegraphics[width=.43\textwidth]{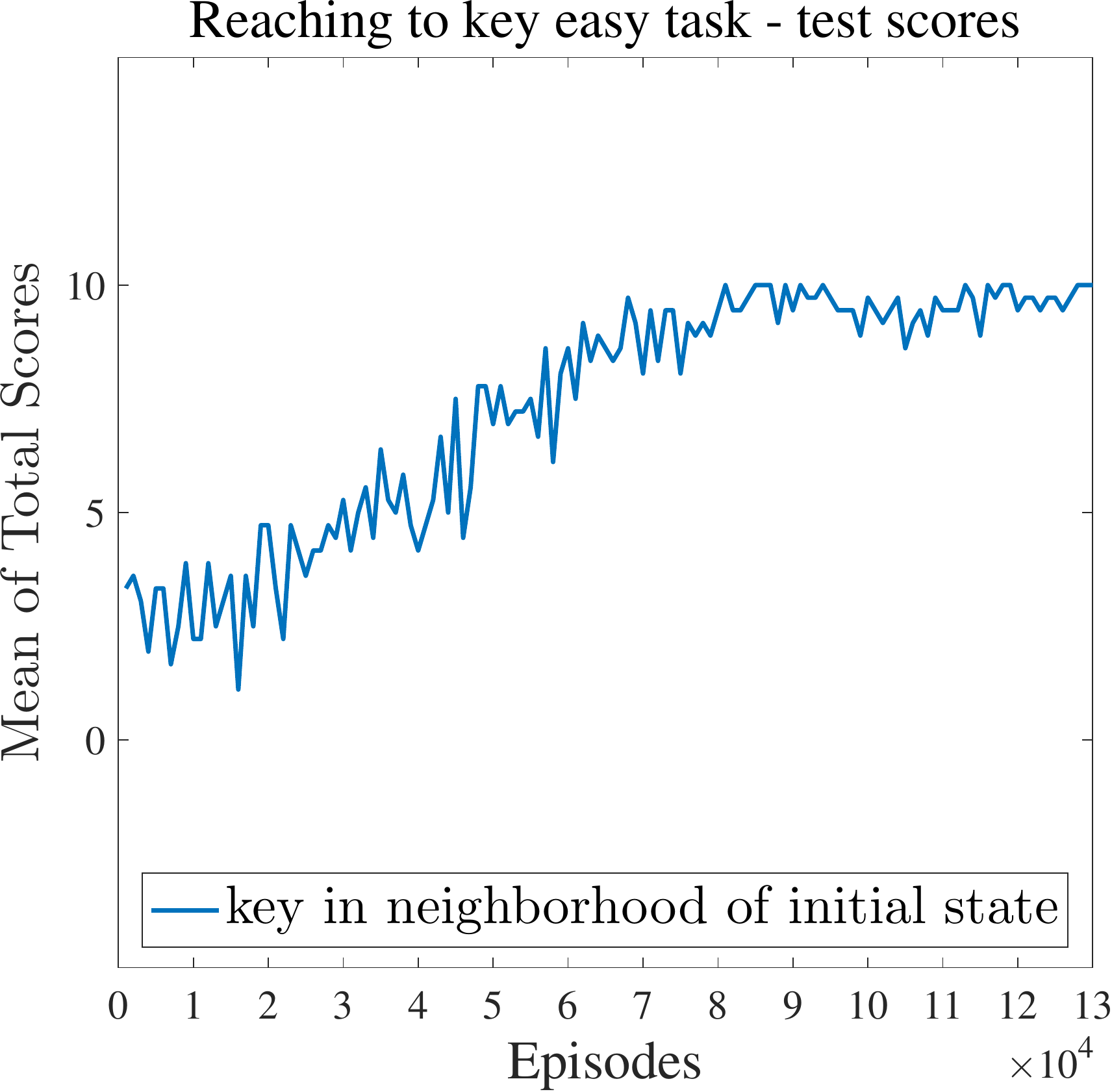} &
		\includegraphics[width=.44\textwidth]{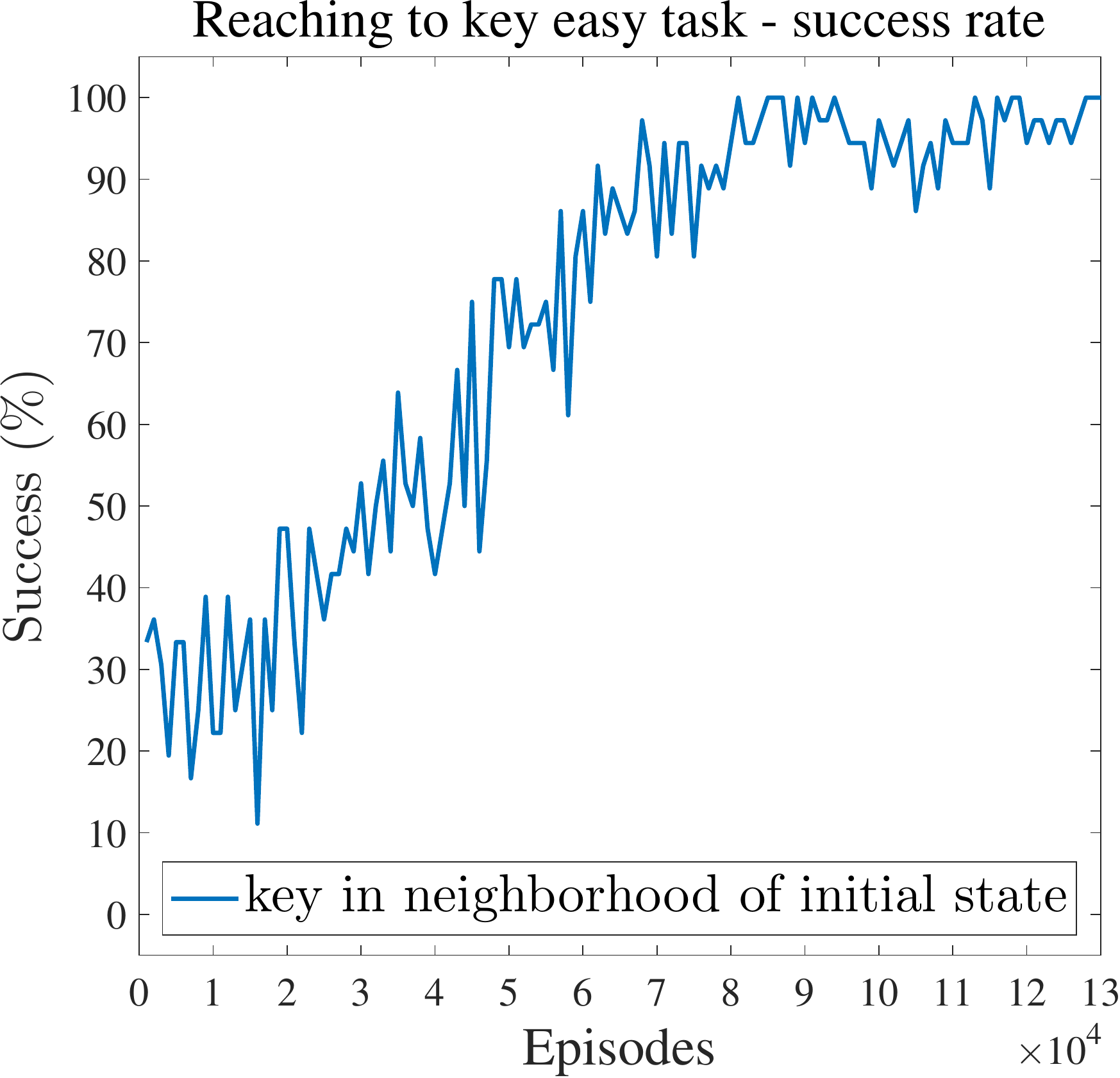}\\
		(c) & (d)
	\end{tabular}\caption{The test results for the task of moving to the key. Top: The key is located in a random location. Bottom: The key is randomly located in the neighborhood of the initial state. The \emph{total scores} are the average of the total reward scores from all possible initial states. The \emph{success rate} is the percentage of the test episodes in which the agent moves to the key.}
	\label{plots:hard-easy-key}
\end{figure*}

\begin{figure*} 
	\centering
	\begin{tabular}{cc} 
		\includegraphics[width=.45\textwidth]{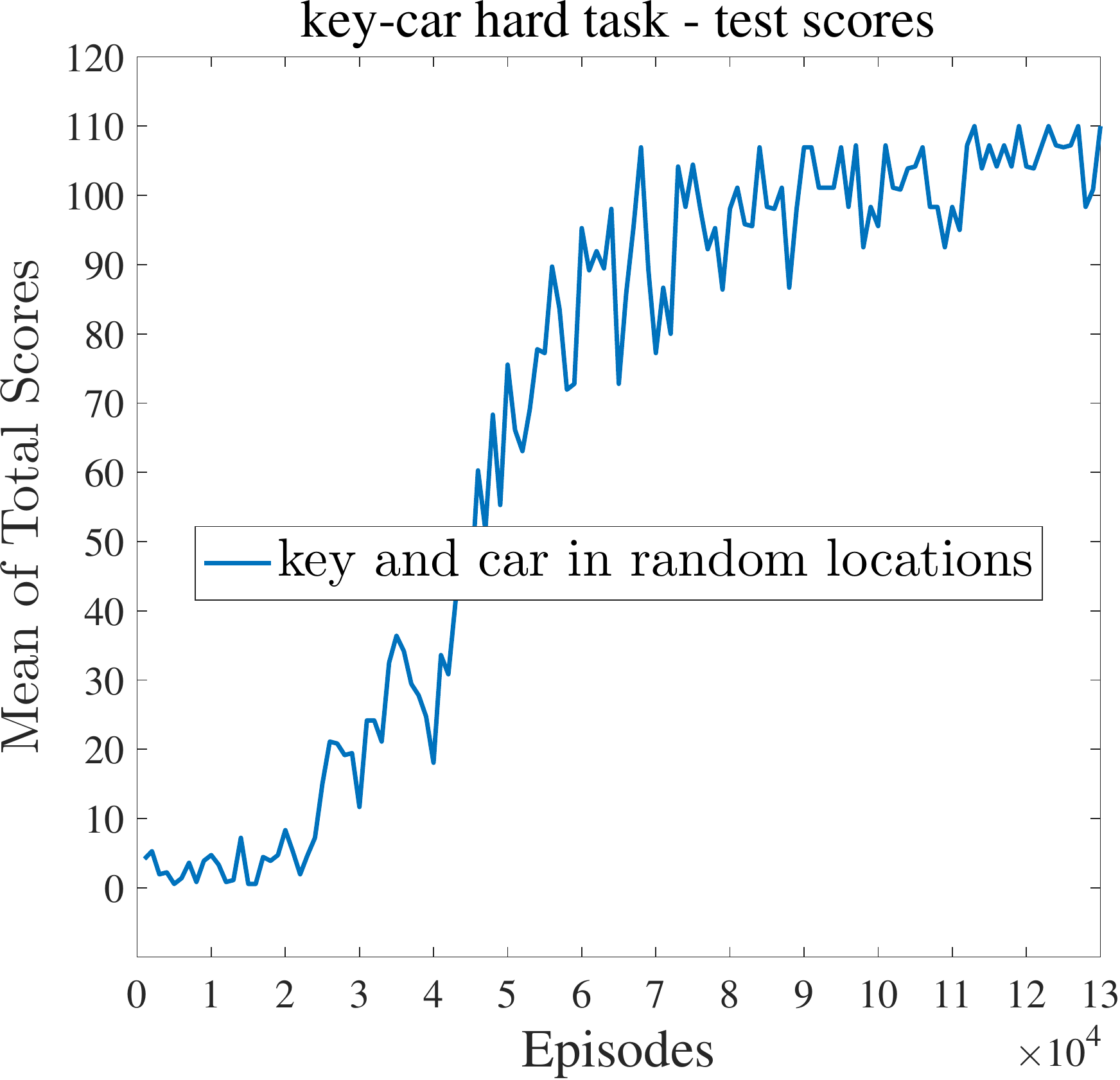} &
		\includegraphics[width=.45\textwidth]{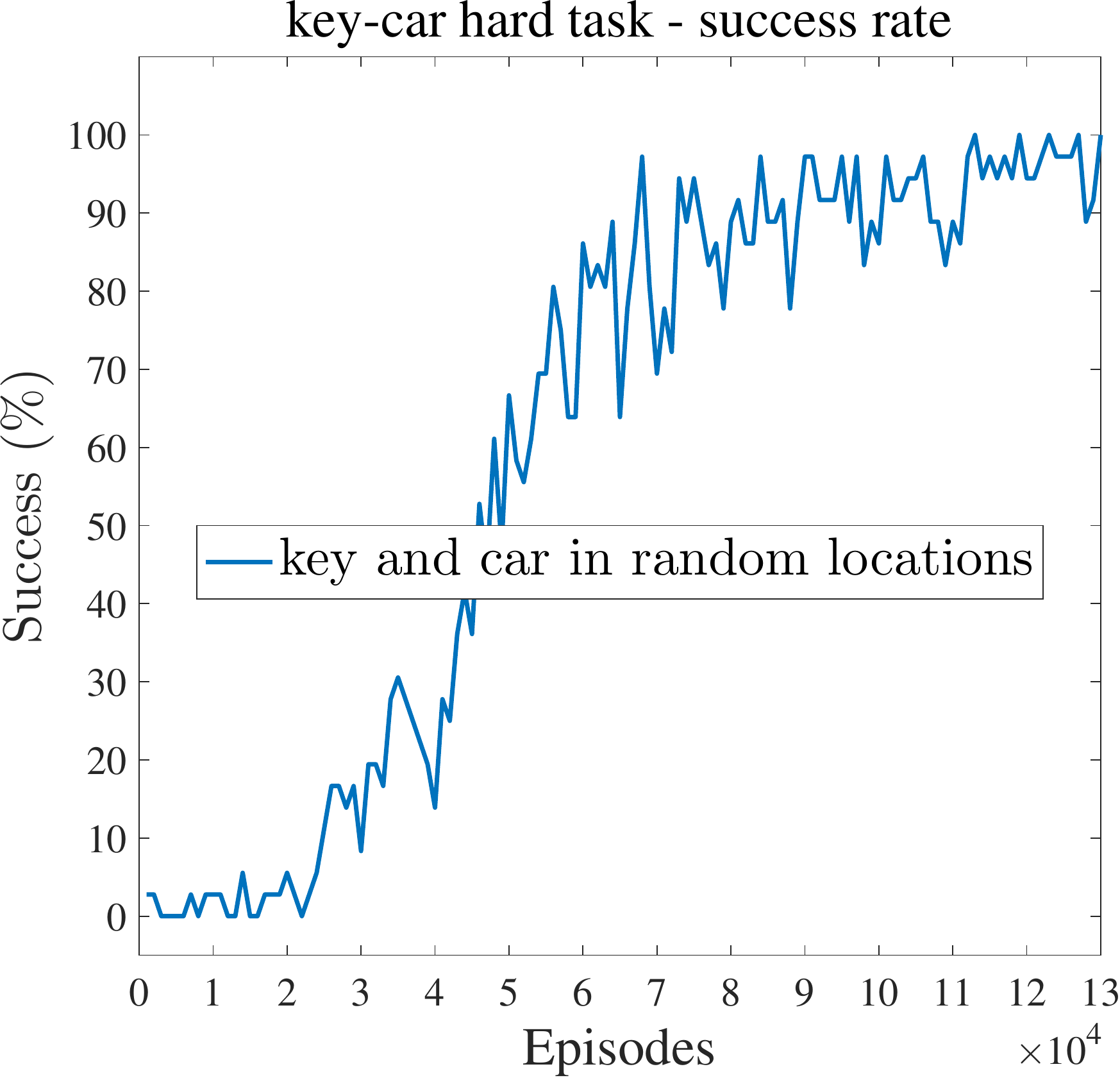}\\
		(a) & (b) \\		
		\includegraphics[width=.45\textwidth]{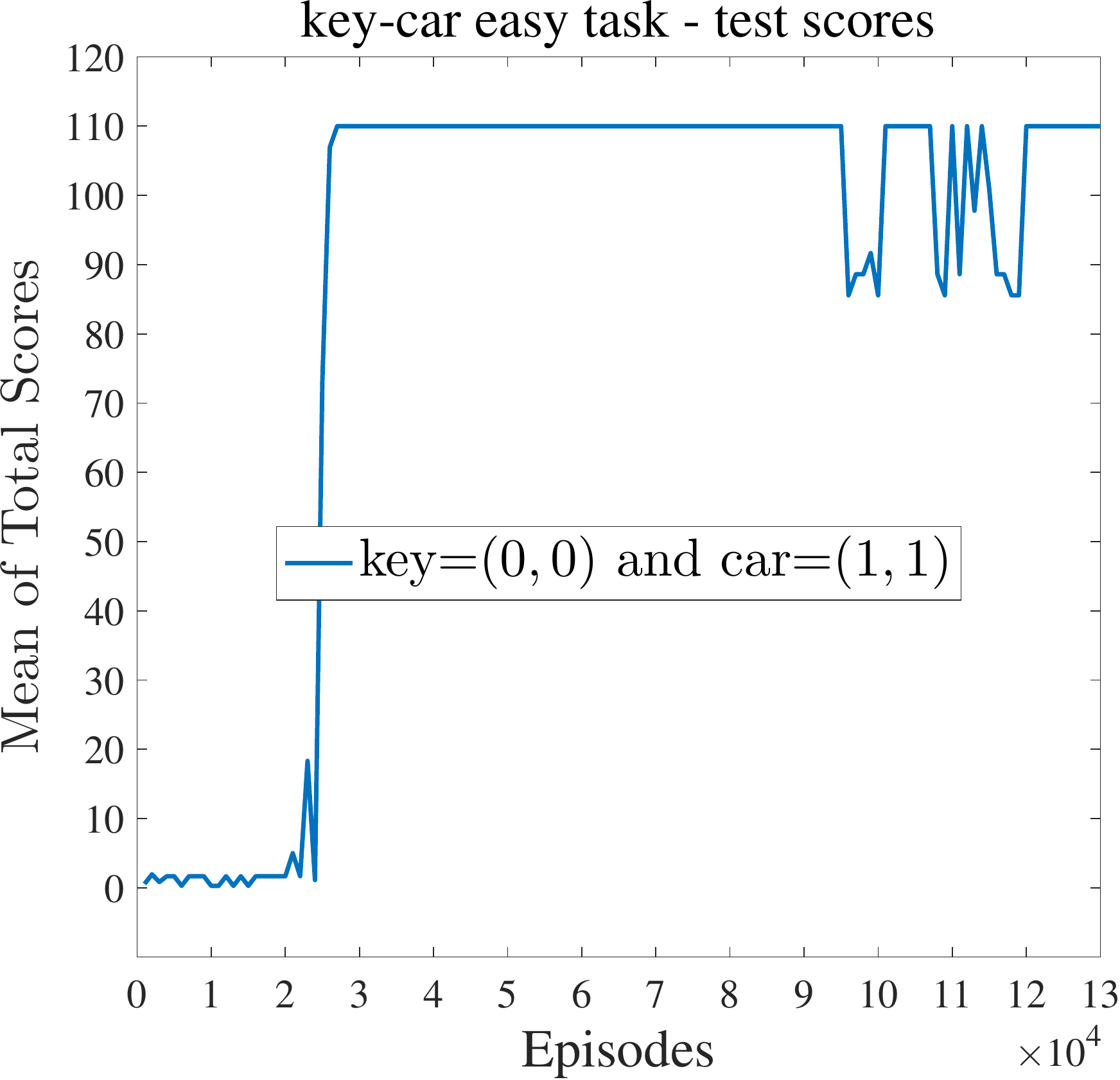} &
		\includegraphics[width=.45\textwidth]{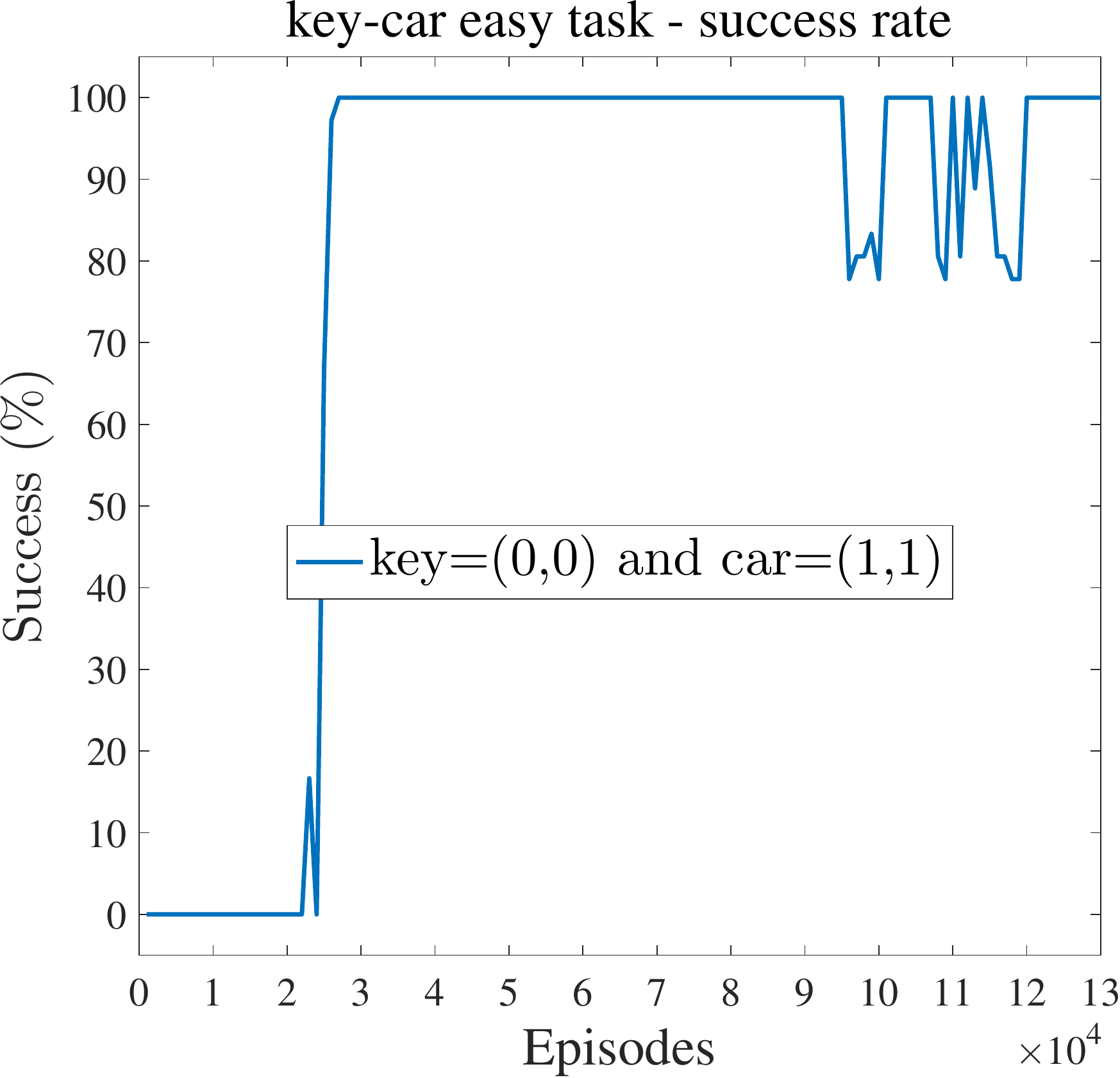}\\
		(c) & (d)
	\end{tabular}\caption{The test results for \emph{key-car} task. Top: \emph{hard placement} --- the key and the car are placed in random locations. Bottom: \emph{easy placement} --- the key is located at (0,0) and the car is located at (1,1). The \emph{total scores} are the average of the total scores form all possible initial states. The \emph{success rate} is the percentage of the test episodes in which the agent successfully moves to the key and then to the car.}
	\label{plots:hard-easy-key-door}
\end{figure*}

The agent was able to move in a two-dimensional grid world environment, containing one key and one car. The agent's location was bounded to be within a 2D Cartesian space of size $[0,1] \times [0,1]$. In the full task, as previously described, the agent received the most reward if it first moved to the key and then moved to the car. In this version of the task, the locations of the key and the car are randomly selected, changing dynamically across training episodes. The agent received no reward or punishment for exploring the space, with the exception of a reward of $r=-2$ if the agent bumped into a wall. Positive rewards were different for different variants of the general task. The agent received complete sensory information of the entire environment (rather than just its own location), including the relative location of the key, the car, and the relative location of the agent, itself in the room.  This additional information was used by the oracle to select the subgoals. Initially, the agent had no semantic knowledge about the objects in the environment. For example, it did not know that, in order to reach the car, it must grab the key first. This situation is illustrated in Figure \ref{grid-world-key-car-env}(b). The agent's internal controller used $q(s,g,a;w)$ to select an action, $a_t$, based on the $\epsilon$-greedy policy. When tested, however, no exploration was allowed, so the policy for a given $g$ can be obtained as 
$\pi_{ag}(s,g) = \arg\max_a q(s,g,a;w)$. At regular intervals during training, we tested the ability of the controller to reach the key, and the car locations in four tasks.
\begin{itemize}
	\item \textbf{Key Task, Hard Placement.} In this simplified version of the task, the agent was trained to move to the key, producing a policy, $\pi_{ag}$, for reaching a randomly located goal $g$ (key). This is illustrated in Figure \ref{grid-world-key-car-env}(a). For each starting $s \in S$, a random goal, $g$, was assigned and the cumulative reward was obtained. We report the average reward scores and the average success percentage in Figure \ref{plots:hard-easy-key} (a) and (b), respectively.
	\item \textbf{Key Task, Easy Placement.}
	This version of the task is the same as the last, except that the goal, $g$, was always randomly placed in a location adjacent to the starting state, $s$. (See Figure \ref{grid-world-key-car-env} (a).) We report the average reward scores and the average success percentage in Figure \ref{plots:hard-easy-key} (c) and (d), respectively.
	\item \textbf{Key-Car Task, Hard Placement.}
	In this version of the task, both the key, $g_{key}$, and the car, $g_{car}$, were randomly placed. The agent received positive reward when the agent moved to the key (+10) and subsequently moved to the car (+100). (See Figure \ref{grid-world-key-car-env} (b).) We report the average scores and the average success percentage in Figure \ref{plots:hard-easy-key-door} (a) and (b), respectively.
	\item \textbf{Key-Car Task, Easy Placement.}
	This version of the task is the same as the last, except that the key was always located at $(0, 0)$, and the car was always located at $(1, 1)$. We report the average reward scores and the average success percentage in Figure \ref{plots:hard-easy-key-door} (c) and (d), respectively.
\end{itemize}

\subsection{Reusing Learned Skills}
\begin{figure} 
	\centering
	\begin{tabular}{cc} 
		\includegraphics[width=.37\textwidth]{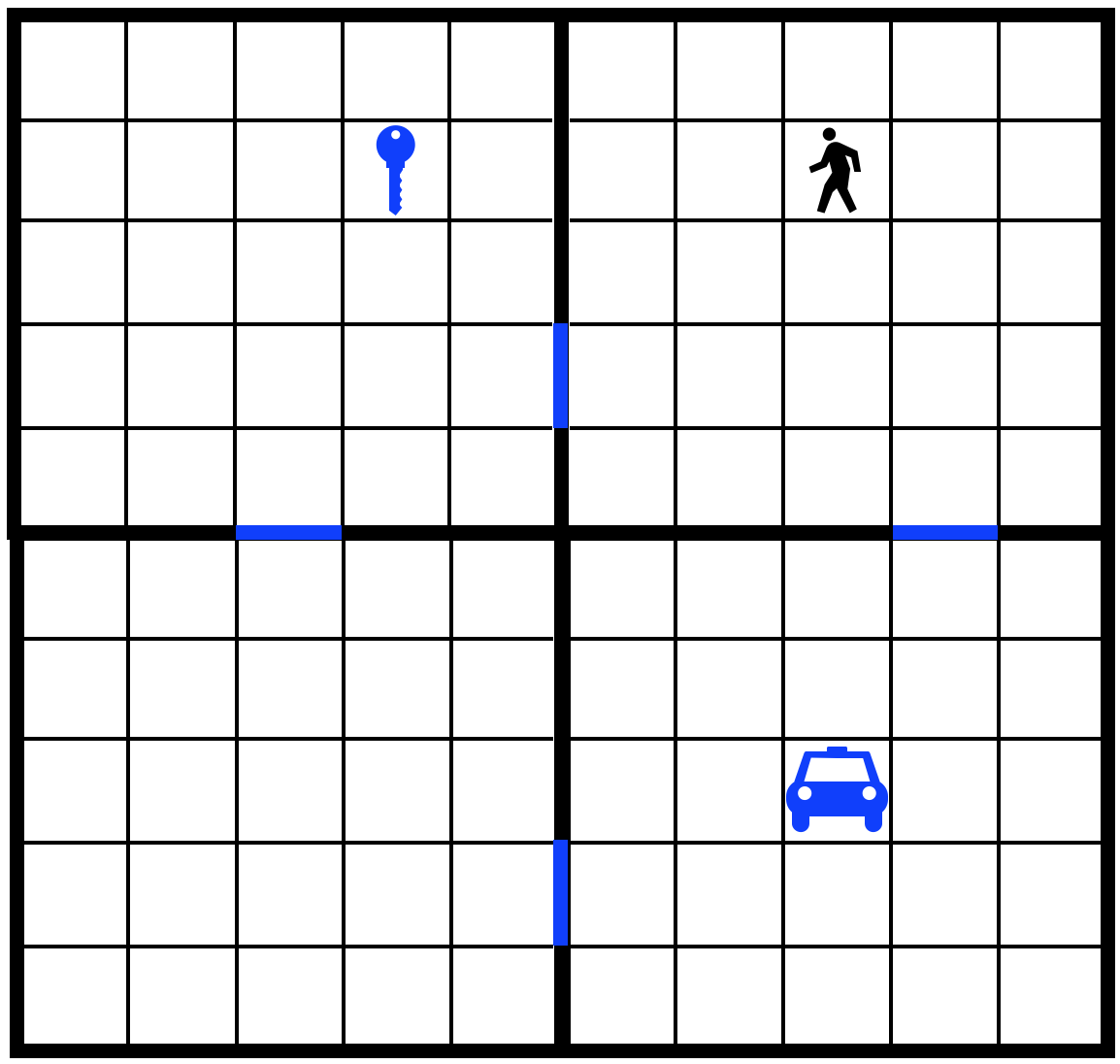} &
		\includegraphics[width=.37\textwidth]{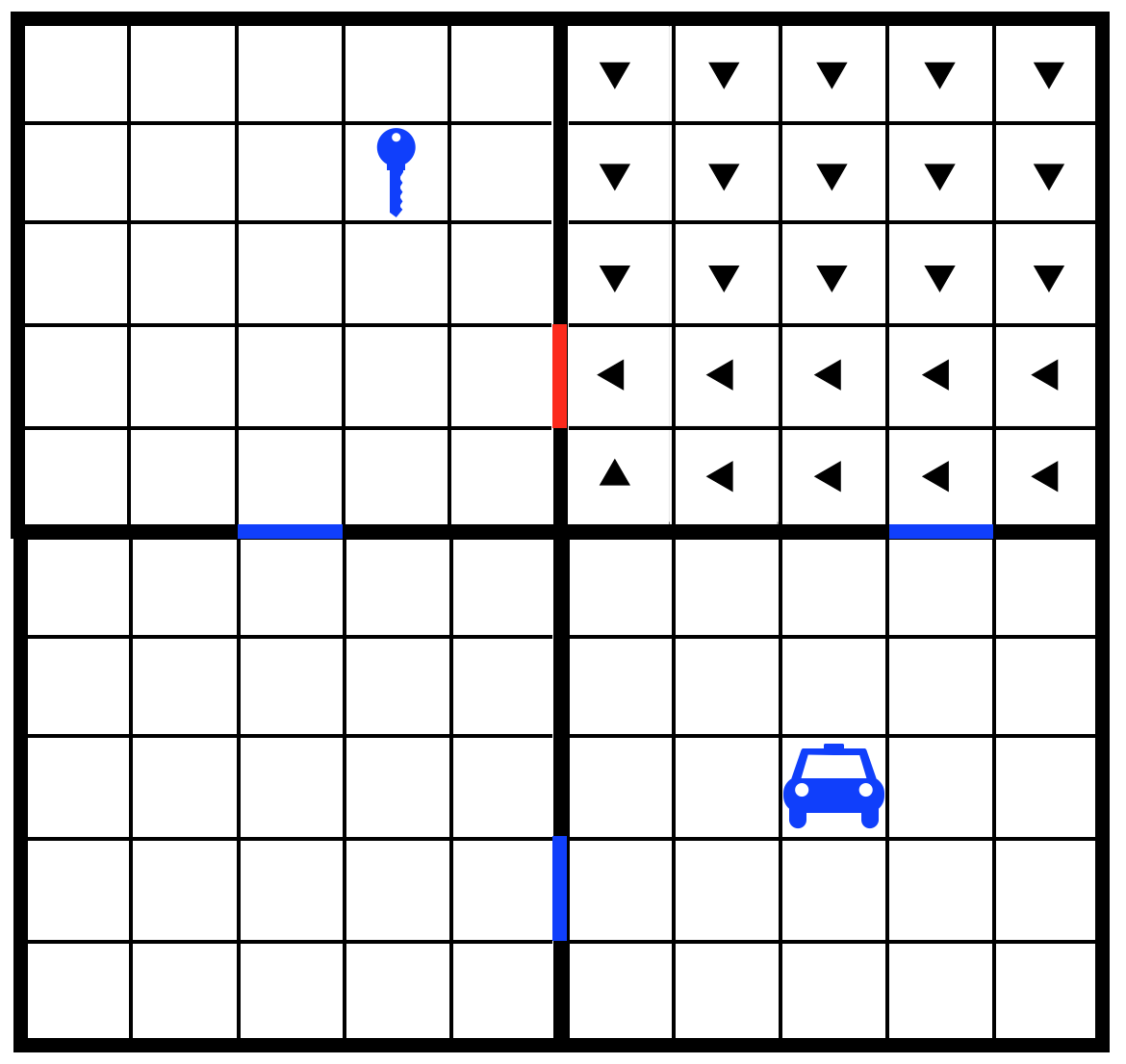} \\
		(a) & (b) \\
		\includegraphics[width=.37\textwidth]{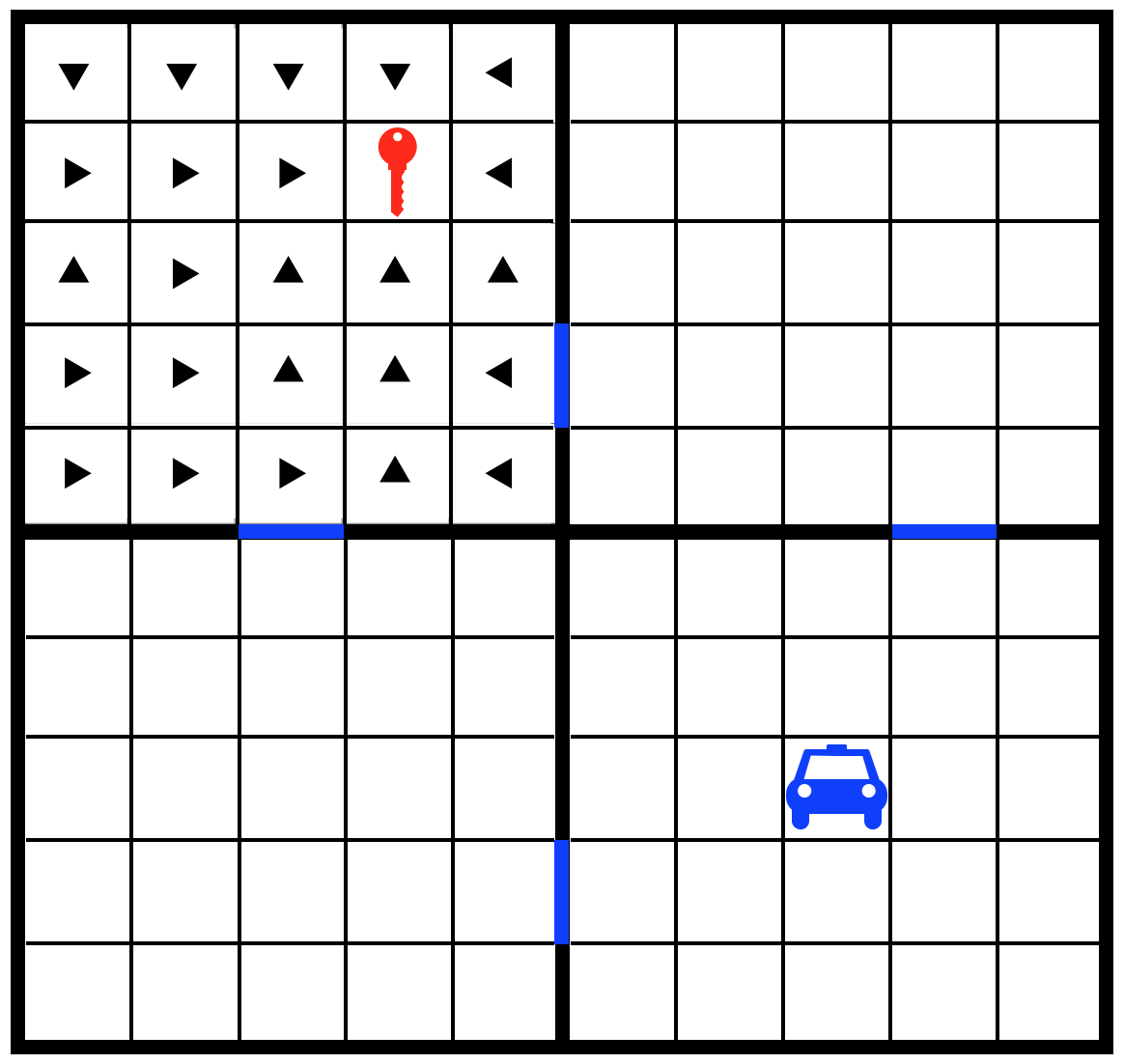} & 
		\includegraphics[width=.37\textwidth]{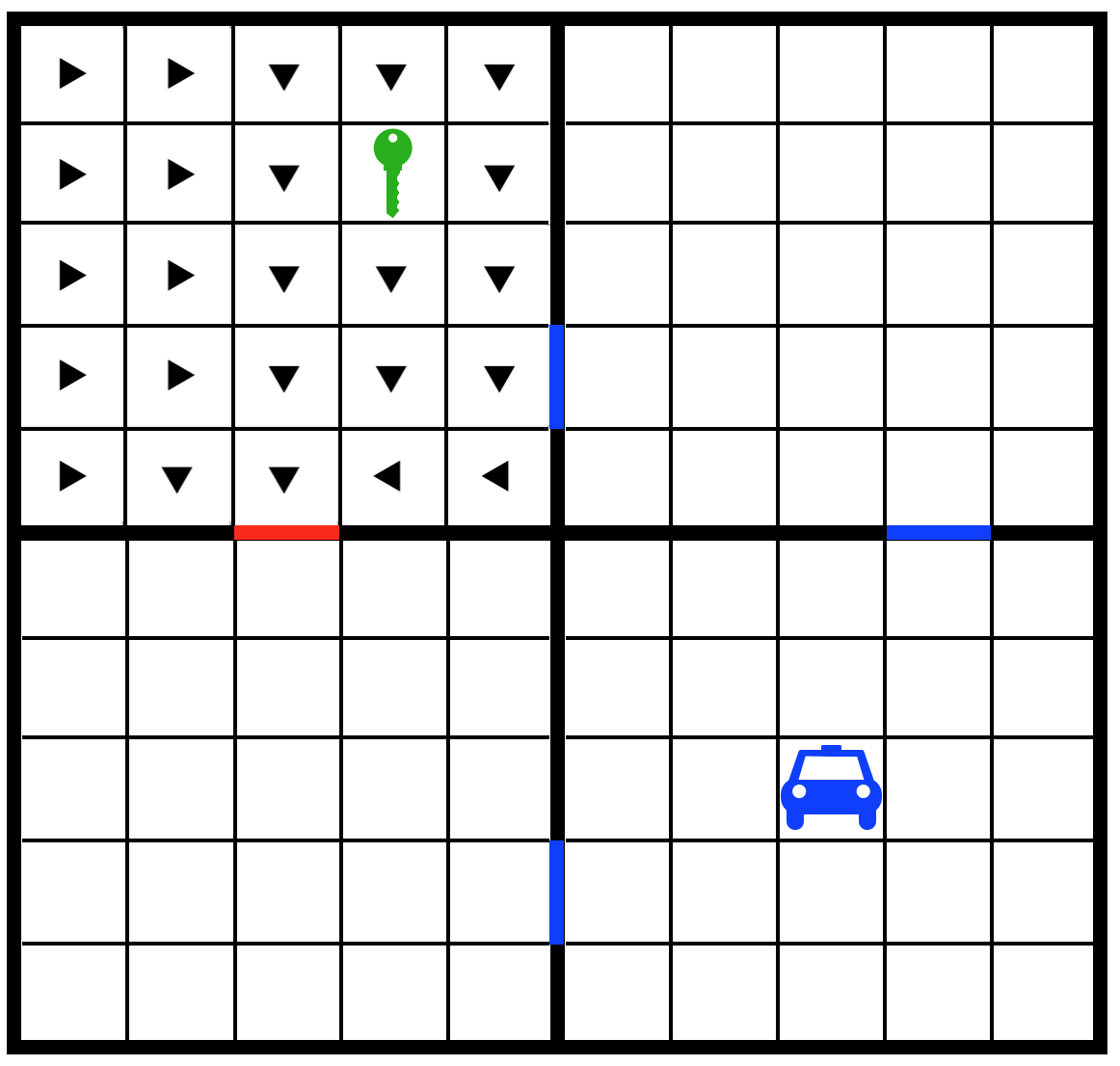} \\
		(c) & (d)\\
		\includegraphics[width=.37\textwidth]{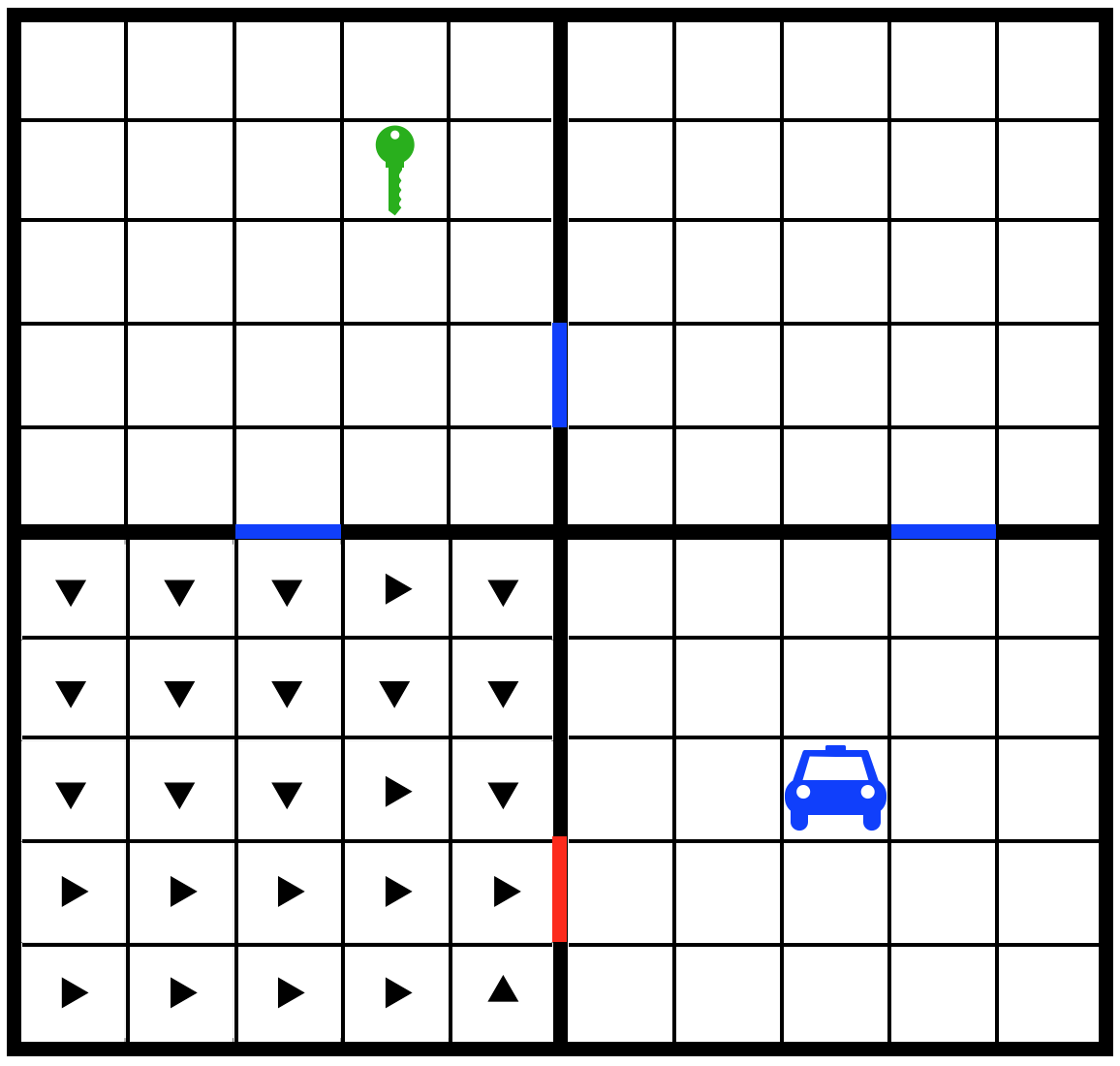} &
		\includegraphics[width=.37\textwidth]{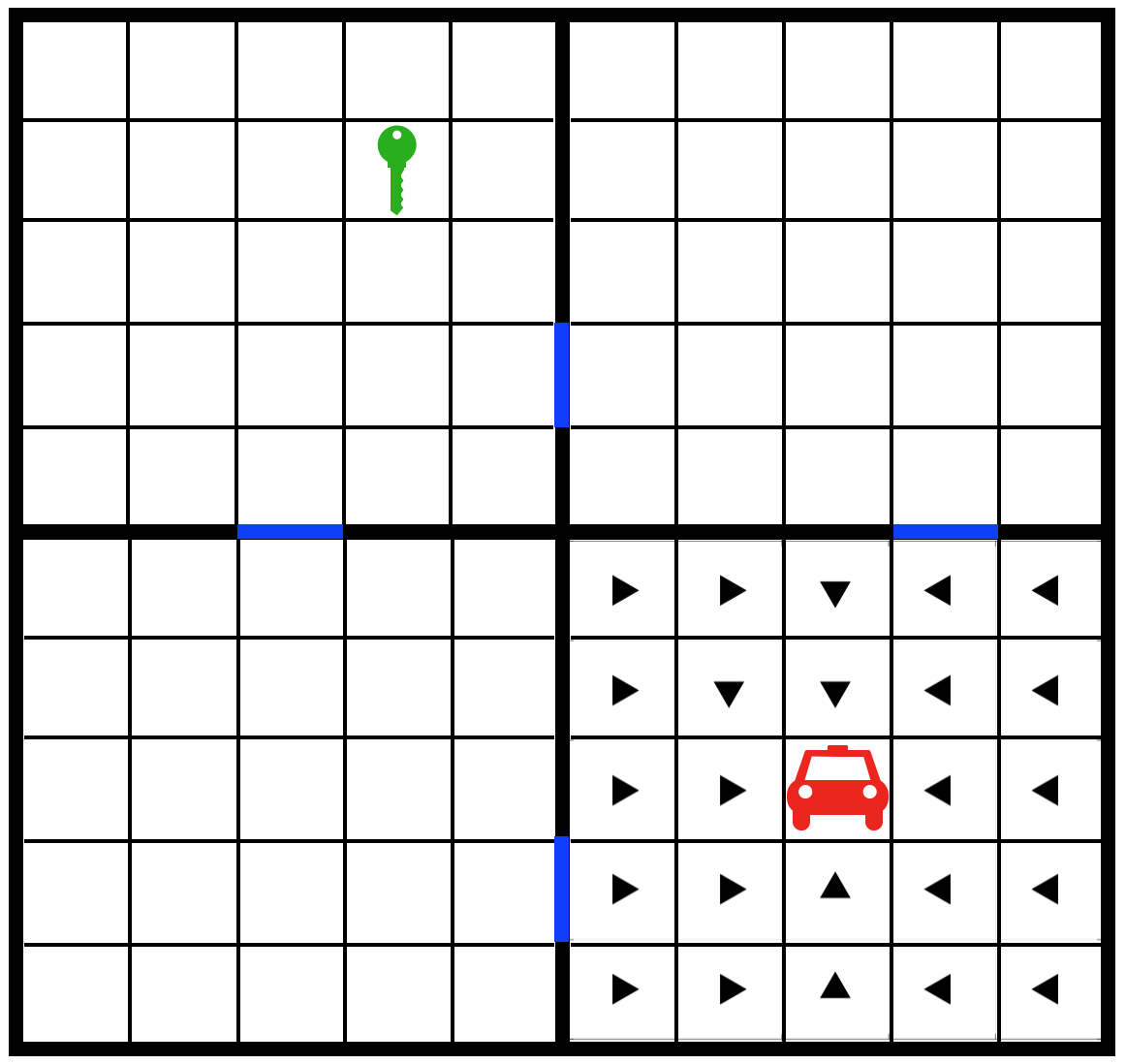} \\
		(e) & (f)
	\end{tabular}\caption{Reusing the navigation skill to solve the rooms task. At each time step, an oracle selected a subgoal for the agent (red locations). The agent with the pretrained navigation skill successfully accomplishes all of the subgoals assigned by the oracle. (a) The starting configuration. (b) Subgoal: doorway between room 1 and 2. (c) Subgoal: moving to the key. (d) Subgoal: doorway between room 2 and 3. (e) Subgoal: doorway between room 3 and 4. (f) Subgoal: moving to the car.}
	\label{plots:policy-4-room-door-key-car}
\end{figure}

\emph{Spacing} the state space, $\mathcal{S}$, through intrinsic motivation enables efficient learning for hierarchical tasks. Consider the rooms task shown in Figure \ref{plots:policy-4-room-door-key-car}(a). For each (slow scale) time step, the meta-controller assigns a subgoal $g \in$\texttt{\{doorways, key, car\}} to the controller. The trained policy can be used to achieve each subgoal, in turn. (See Figure \ref{plots:policy-4-room-door-key-car} (b) to (f).) The controller is trained to solve the navigation task (in a single room gridworld) for any given goal $g$. The intrinsic motivation makes solving the rooms task easier since the agent can adapt the policy to the given goal $g$. It is important to note that we made an assumption that an oracle in the meta-controller provides the transformation from the state space to the input state for the network. This assumption is discarded in later experiments in Section \ref{sec:hrl:exp} 

\section{Unsupervised Subgoal Discovery}
The performance of the meta-controller/controller framework depends
critically on selecting good candidate subgoals for the
meta-controller to consider. 

What is a subgoal? In our framework, a subgoal is a state, or a set of
related states, that satisfies at least one of these conditions:
\begin{enumerate}
	\item It is close (in terms of actions) to a rewarding state. For
	example, in the rooms task in Figure \ref{fig:rooms}(a), the key
	and lock are rewarding states.
	\item It represents a set of states, at least some of which tend to be
	along a state transition path to a rewarding state. 
\end{enumerate}
For example, in the rooms task, the red room, as illustrated in Figure \ref{fig:rooms} (a), should be visited to move
from the purple room to the blue room in order to pick up the
key. Thus any state in the red room is a reasonably good subgoal for
an agent currently in the purple room. Similarly, the states in the
blue room are all reasonably good subgoals for an agent currently in
the red room. The doorways between rooms can also be considered as
good subgoals, since entering these states allows for the transition
to a set of states that may be closer to rewarding states.

Our strategy involves leveraging the set of recent transition
experiences that must be recorded for value function learning,
regardless. Unsupervised learning methods applied to sets of
experiences can be used to identify sets of states that may be good
subgoal candidates. We focus specifically on two kinds of analysis
that can be performed on the set of transition experiences. We
hypothesize that good subgoals might be found by (1) attending to the
states associated with \emph{anomalous} transition experiences and (2)
clustering experiences based on a similarity measure and collecting
the set of associated states into a potential subgoal. Thus, our
proposed method merges \emph{anomaly (outlier)
	detection} with the $K$-means clustering of experiences. The unsupervised subgoal discovery method is summarized in Algorithm \ref{Algo:USD}.
\begin{algorithm}
	\begin{algorithmic}
		\For{each $e=(s,a,r,s')$ stored in $\mathcal{D}$}
		\If{experience $e$ is an outlier (anomaly)}
		\State Store $s'$ to the subgoals set $\mathcal{G}$
		\State Remove $e$ from $\mathcal{D}$ 
		\EndIf
		\EndFor
		\State Fit a $K$-means Clustering Algorithm on $\mathcal{D}$ using previous centroids as initial points   
		\State Store the updated centroids to the subgoals set $\mathcal{G}$					
	\end{algorithmic}
	\caption{Unsupervised Subgoal Discovery Algorithm}
	\label{Algo:USD}
\end{algorithm}

\subsection{Anomaly Detection}
The anomaly (outlier) detection process identifies states associated
with experiences that differ significantly from the others. In the
context of subgoal discovery, a relevant anomalous experience would be
one that includes a substantial positive reward in an environment in
which reward is sparse. We propose that the states associated with
these experiences make for good candidate subgoals. For example, in
the rooms task, transitions that arrive at the key or the lock are
quite dissimilar to most transitions, due to the large positive reward
that is received at that point. 

Since the goal of RL is maximizing accumulated (discounted) reward,
these anomalous experiences, involving large rewards, are ideal as
subgoal candidates. {(Experiences involving large negative rewards
	are also anomalous, but make for poor subgoals. As long as these sorts
	of anomalies do not greatly outnumber others, we expect that the
	meta-controller can efficiently learn to avoid poor subgoal choices.)}
Large changes in state features can also be marked as anomalous. In
some computer games, like \emph{Montezuma's Revenge}, each screen
represents a room, and the screen changes quickly when the agent moves
from one room to another. This produces a large distance between two
consecutive states. Such a transition can be recognized simply by the
large instantaneous change in state features, marking the associated
states as reasonable candidate subgoals. There is a large literature
on anomaly detection \citep{Hodge2004-anomaly-survey}, in general,
offering methods for applying this insight. Heuristic meta-parameter
thresholds can be used to identify dissimilarities that warrant
special attention, or unsupervised machine learning methods can be
used to model the joint probability distribution of state variables,
with low probability states seen as anomalous.

\subsection{K-Means Clustering}
The idea behind using a clustering algorithm is ``spatial'' state
space abstraction and dimensionality reduction with regard to the
internal representations of states. If a collection of transition
experiences are very similar to each other, this might suggest that
the associated states are all roughly equally good as subgoals. Thus,
rather than considering all of those states, the learning process
might be made faster by considering a representative state (or smaller
set of states), such as the centroid of a cluster, as a
subgoal. Furthermore, using a simple clustering technique like
$K$-means clustering to find a small number of centroids in the space
of experiences is likely to produce centroid subgoals that are
dissimilar from each other. Since rewards are sparse, this
dissimilarity will be dominated by state features. For example, in the
rooms task, the centroids of $K$-means clusters, with $K=4$, lie
close to the geometric centers of the rooms, with the states within
each room coming to belong to the corresponding subgoal's cluster. In
this way, the clustering of transition experiences can approximately
produce a coarser representation of state space, in this case
replacing the fine grained ``grid square location'' with the coarser
``room location''.

\subsection{Mathematical intuition}

The value of a state, $V_{\pi}(s)$, is defined as the expected future rewards, following a policy $\pi$ 
\begin{align}
V_{\pi}(s) \triangleq \mathbf{E} \big[ \sum_{t=0}^{\mathcal{T}} \gamma^{t} r_t  |  s,\pi  \big],
\end{align}
where $\mathcal{T}$ is a termination time, and $\gamma<1$. In the  model-free HRL framework, $V_{\pi}(s)$ can be
approximated by a sequence of values of the meta-controller's value function for one subgoal after another
\begin{align}
V(s) \approx Q(s,g_1) + \gamma^{T_1} Q(g_1,g_2) + \gamma^{T_1+T_2}
Q(g_2,g_3) + \dots 
\end{align}
where $T_i \leq T$
is effective number of controller steps to accomplish subgoal
$g_i$ (note that $\mathcal{T} = T_1 + T_2 + \dots$). We assume that the controller has learned a good policy through the process of intrinsic motivation learning. Since the rewards are sparse, the value of an state close to an
anomalous subgoal is roughly equal to the immediate reward $r$ (since the future rewards vanish for a small discount factor $\gamma$). The $K$-means clustering algorithm partitions states space
into $K$ regions. The value of states in a cluster close to an anomalous
subgoal $g$ is approximately $\gamma^{T_1}r$, since it takes $T_1$ steps from states in this region to arrive the rewarding state, and obtain the reward $r$. The clustering algorithm takes into consideration the distance between experiences that do not contain anomalous ones. Therefore, the states in a cluster have similar values, because, for each state in a cluster, it takes approximately the same number of steps to reach a rewarding state (an anomalous subgoal).    

\section{A Unified Model-Free HRL Framework}
\label{sec:unified-hrl}
In this section, we introduce a unified method for model-free HRL, so that all three HRL subproblems can be solved jointly. Our intuition, shared with other researchers, is that hierarchies of abstraction will be critical for successfully solving problems with sparse delayed feedback. To be successful, the agent should represent knowledge at
multiple levels of spatial and temporal abstraction. Appropriate
abstraction can be had by identifying a relatively small set of states
that are likely to be useful as \emph{subgoals} and jointly learning
the corresponding skills of achieving these subgoals, using intrinsic
motivation.

\begin{figure*}
	\centering
	\begin{tabular}{cc}
		\includegraphics[width=0.32\textwidth]{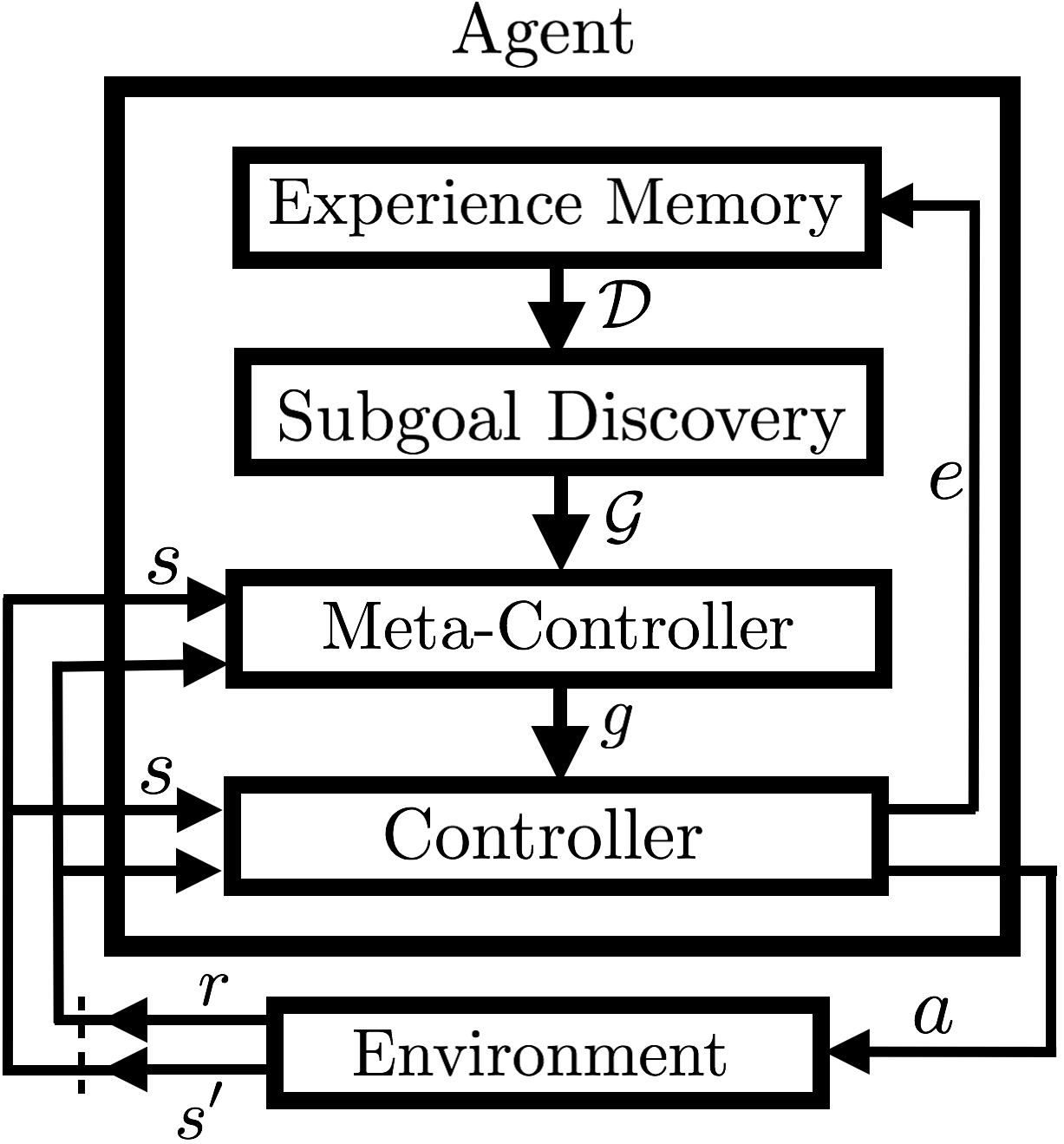} &
		\includegraphics[width=0.6\textwidth]{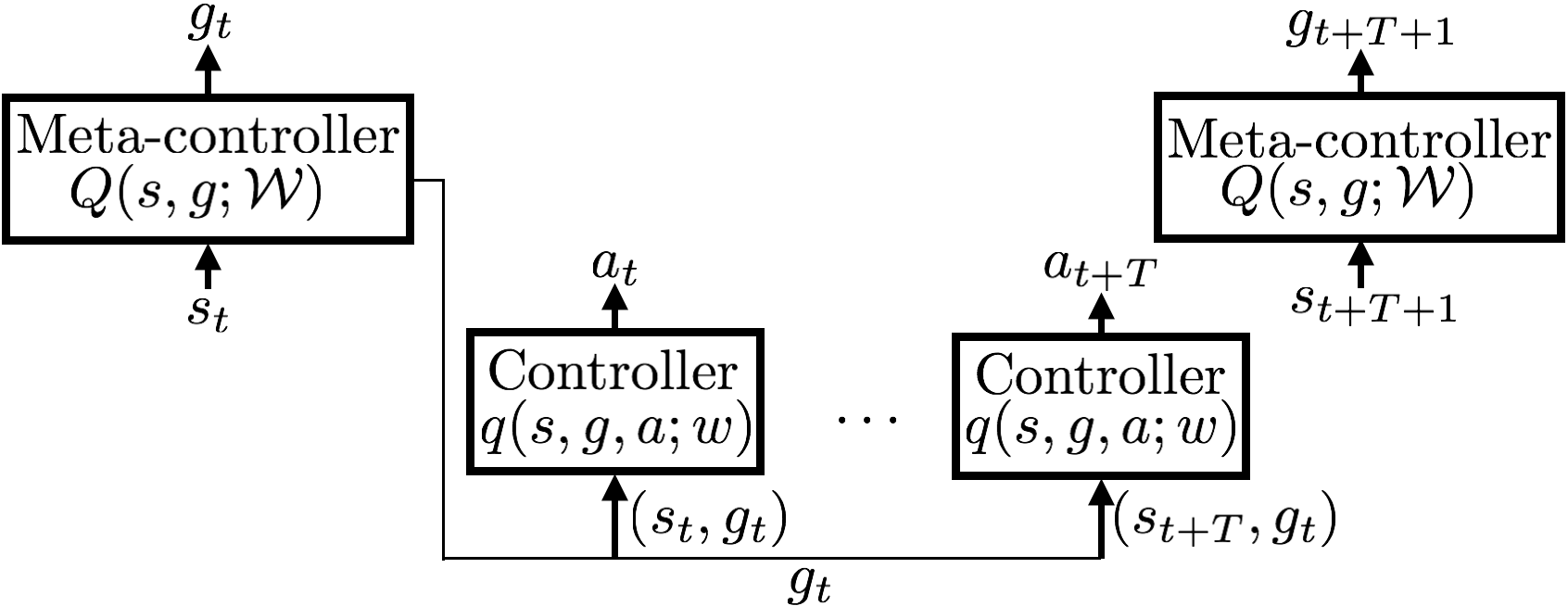}\\
		(a) & (b)
	\end{tabular}
	\caption{(a) The information flow in the unified Model-Free Hierarchical Reinforcement Learning Framework. (b) Temporal abstraction in the meta-controller/controller framework.}
	\label{fig:unified-hrl-framework}
\end{figure*}

Inspired by Kulkarni et al. (\citeyear{Kulkarni:2016:Meta-Controller}), we start by using two levels of hierarchy (Figure \ref{fig:unified-hrl-framework}). The more abstract level of this hierarchy is managed by a \emph{meta-controller} which guides the action selection processes of the lower level \emph{controller}. Separate value functions are learned for the meta-controller and the controller as shown in Figure \ref{fig:unified-hrl-framework}(b).

The conceptual components of HRL --- temporal abstraction, intrinsic motivation, and unsupervised subgoal discovery --- can be unified into a single model-free HRL framework. The major components of this framework, and the information flow between these components, are schematically
displayed in Figure \ref{fig:unified-hrl-framework} (a). At time $t$,
the meta-controller observes the state, $s=s_t$, from the environment and
chooses a subgoal, $g=g_t$, either from the discovered subgoals or from a
random set of states (to promote exploration). The controller receives
an input tuple, $(s,g)$, and is expected to learn to implement a
subpolicy, $\pi({a | s,g})$, that solves the \emph{subtask} of
reaching from $s$ to $g$. The controller selects an action, $a$, based
on its policy, in our case directly derived from its value function,
$q(s,g,a;w)$. After one step, the environment updates the state to
$s'$ and sends a reward $r$. The agent's experience $(s,a,s',r)$ is stored in the experience memory, $\mathcal{D}$. The intrinsic transition experience
$(s,g,a,\tilde{r},s')$ is stored in the controller's experience memory, $\mathcal{D}_{1}$. If the internal critic detects that the
resulting state, $s'$, is the current goal, $g$, the experience
$(s_t,g,G,s_{t'})$ is stored in the meta-controller experience memory,
$\mathcal{D}_{2}$, where $s_t$ is the state that prompted the
selection of the current subgoal, and $s_{t'}=s_{t+T}$ is the state when the meta-controller assigns the next subgoal, $g'=g_{t'}$. The experience memory sets are typically used to train the value function approximators for the meta-controller and the controller by sampling a random minibatch of recent experiences. The unsupervised subgoal discovery mechanism exploits the underlying structure in the experience memory $\mathcal{D}$ using unsupervised anomaly detection and $K$-means clustering. A detailed description of the unified representation learning in our model-free HRL framework is outlined in Algorithm \ref{Algo:unified-model-free-hrl}.
\begin{algorithm}[hbt!]
	\begin{algorithmic}  
		\State Pretrain controller using Algorithm \ref{Algo:intrinsic-motivation} on a set of random subgoals $\mathcal{G}'$ 
		\State Initialize experience memories $\mathcal{D}$, $\mathcal{D}_1$ and $\mathcal{D}_2$
		\State Walk controller for $M'$ episodes on random subgoals $\mathcal{G}'$, and store $(s,a,s',r)$ to $\mathcal{D}$
		\State Run Unsupervised Subgoal Discovery on $\mathcal{D}$ to initialize $\mathcal{G}$
		\For{ episode $=1,\dots,M$} 
		\State Initialize state $s_0 \in \mathcal{S}$, $s\gets s_0$
		\State $G \gets 0$
		\State $g \gets$\texttt{EPSILON-GREEDY}$(Q(s,\mathcal{G};\mathcal{W}),\epsilon_2)$
		\Repeat{ for each step $t = 1,\dots,T$} 					
		\State compute $q(s,g,a;w)$
		\State $a\gets$\texttt{EPSILON-GREEDY}$(q(s,g,\mathcal{A};w),\epsilon_1)$
		\State Take action $a$, observe $s'$ and external reward $r$ 
		\State Compute intrinsic reward $\tilde{r}$ from internal critic
		\State Store controller's intrinsic experience, $(s,g,a,\tilde{r},s')$ to $\mathcal{D}_1$	
		\State Store agent's transition experience, $(s,a,r,s')$ to $\mathcal{D}$		
		\State Sample $J_1 \subset \mathcal{D}_1$ and compute $\nabla L$
		\State Update controller's parameters, $\quad w \gets w - \alpha_1 \nabla L$
		\State Sample $J_2 \subset \mathcal{D}_2$ and compute $\nabla \mathcal{L}$
		\State  Update meta-controller's parameters, $\quad \mathcal{W} \gets \mathcal{W} - \alpha_2 \nabla \mathcal{L}$ 
		\State $s \gets s',\quad G \gets G + r$
		\State Decay exploration rate of controller $\epsilon_1$
		\If{experience $e$ is an outlier (anomaly)}
		\State Store $s'$ to the subgoals set $\mathcal{G}$
		\State Remove $e$ from $\mathcal{D}$ 
		\EndIf
		\Until{$s$ is terminal or subgoal $g$ is attained}			
		\State Decay exploration rate of meta-controller $\epsilon_2$
		\State Store meta-controller's experience, $(s_0,g,G,s')$ to $\mathcal{D}_2$
		\State Fit a $K$-means clustering on $\mathcal{D}$ every $N$ step to update centroids of $\mathcal{G}$	
		\EndFor					
	\end{algorithmic}
	\caption{Unified Model-Free HRL Algorithm}
	\label{Algo:unified-model-free-hrl}
\end{algorithm}

\section{Experiments on Unified HRL Framework}
\label{sec:hrl:exp}
We conducted simulation experiments in order to investigate the
ability of our unsupervised subgoal discovery method to discover
useful subgoals, as well as the efficiency of our unified model-free
hierarchical reinforcement learning framework. The simulations were
conducted in two environments with sparse delayed feedback: a variant
of the rooms task, shown in Figure \ref{fig:rooms}(a), and the
``Montezuma's Revenge'' game, shown in Figure \ref{fig:montezuma}(a). 

All codes was implemented in the Python using Pytorch, NumPy, Opencv, and SciPy libraries and is available at \url{https://github.com/root-master/unified-hrl}.

\subsection{4-Room Task with Key and Lock}
\begin{figure*}[hbt!] 
	\centering
	\begin{tabular}{cc}
		\includegraphics[width=0.47\textwidth]{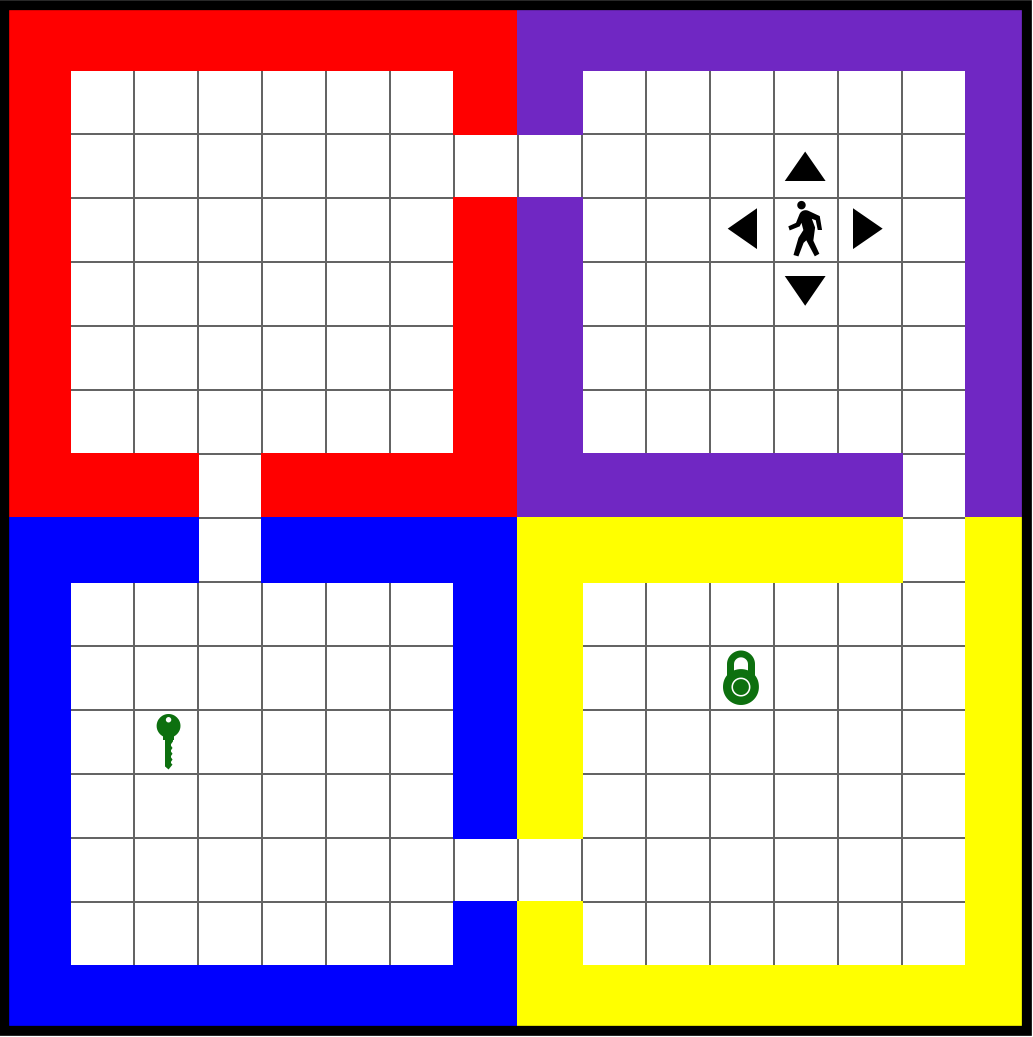} &
		\includegraphics[width=0.47\textwidth]{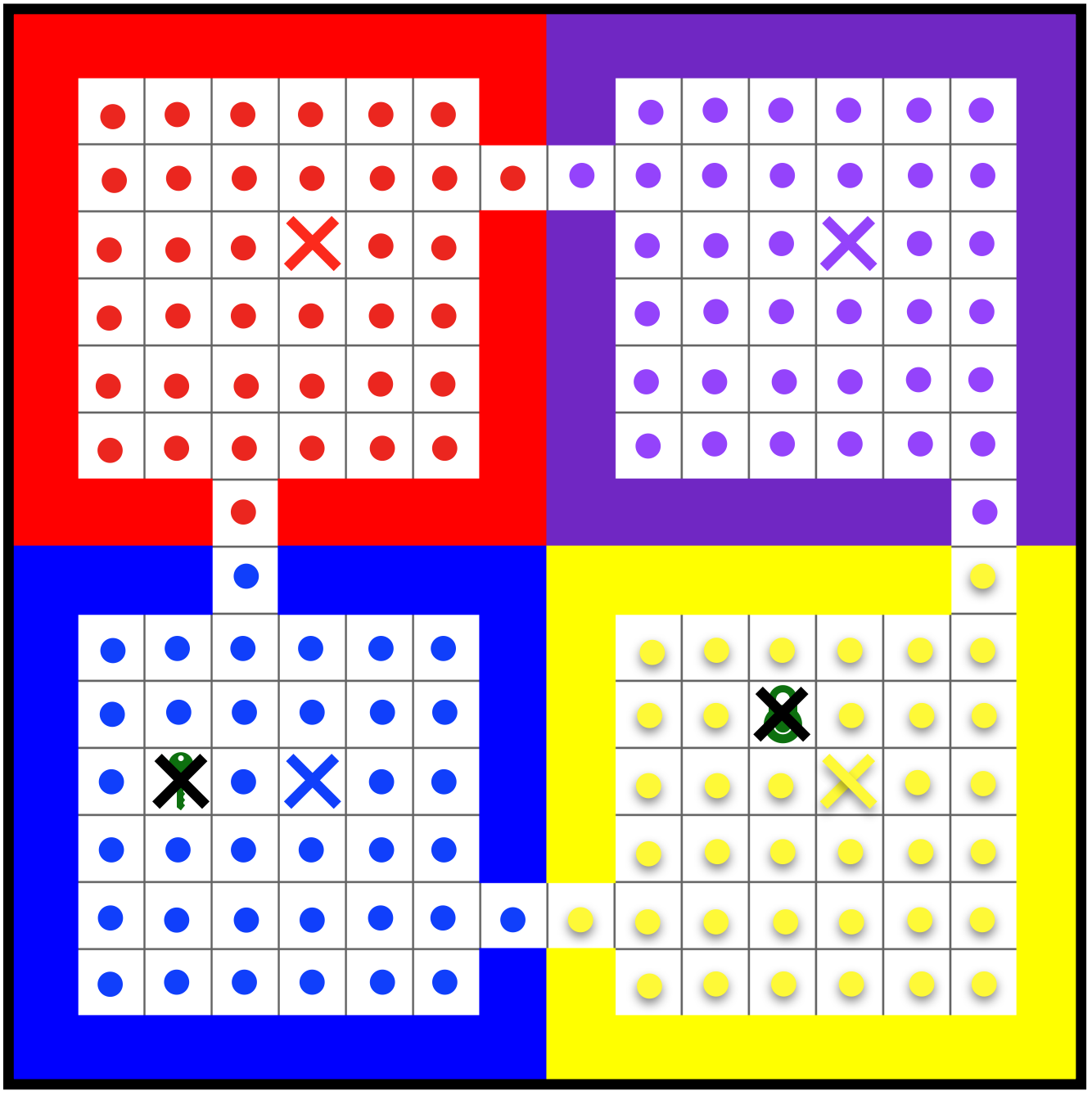}\\
		(a) & (b) \vspace{0.5cm}\\
		\includegraphics[width=0.47\textwidth]{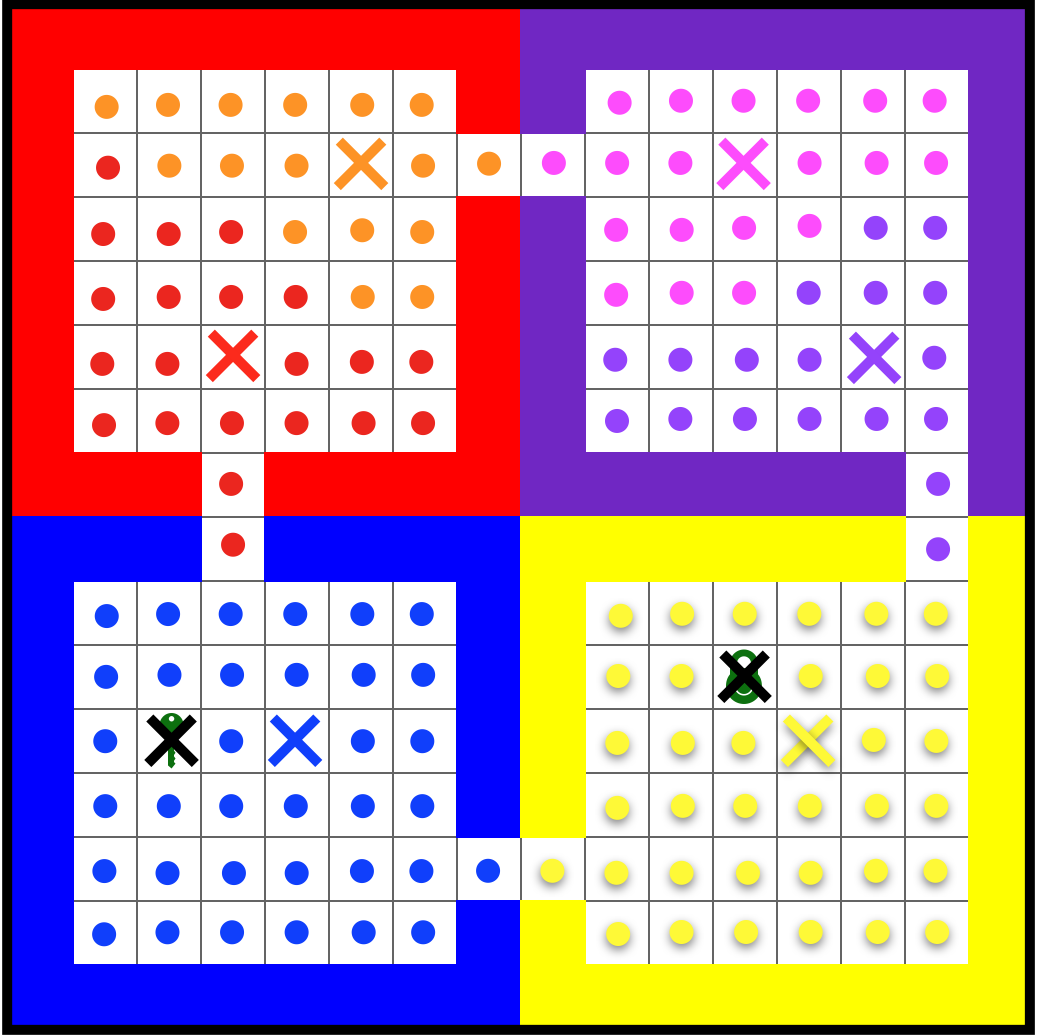} &
		\includegraphics[width=0.47\textwidth]{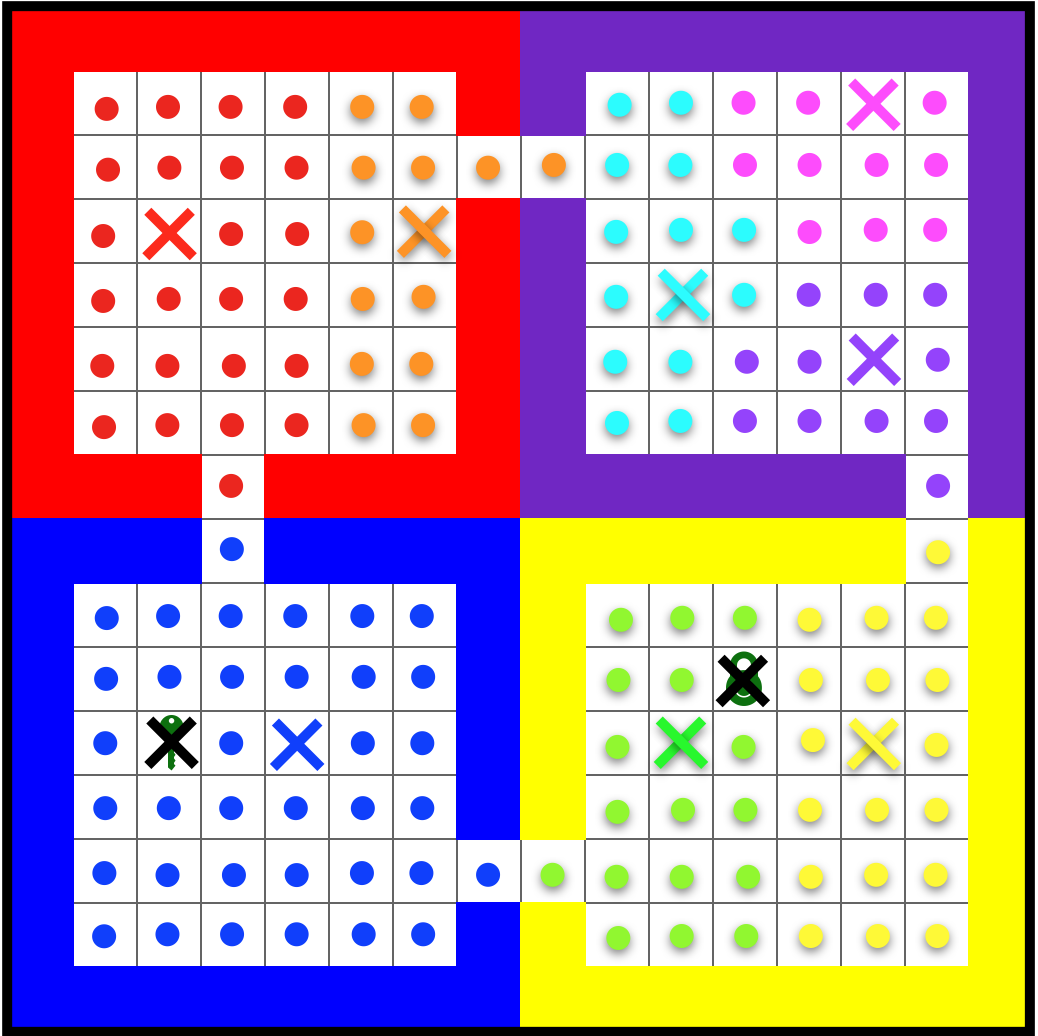}\\
		(c) & (d)  
	\end{tabular}
	\caption{(a) The \emph{4-room} task with a {key} and a {lock}. (b) The results of the unsupervised subgoal discovery algorithm with \emph{anomalies} marked with black Xs and \emph{centroids} with colored ones. The number of clusters in $K$-means algorithm was set to $K=4$. (c) The result of the unsupervised subgoal discovery for $K=6$.  (d) The results of the unsupervised subgoal discovery for $K=8$.}
	\label{fig:rooms}
\end{figure*}

Consider the task of navigation in the \emph{4-room environment with a
	key and a lock}, as shown in Figure \ref{fig:rooms}(a). This is the same task that was explored in earlier parts of this paper. While this task
was inspired by the \emph{rooms} environment introduced by Sutton, et
al. (\citeyear{Sutton:1999:Option}), it is much more complex. As before, the
agent not only needs to learn how to navigate form any arbitrary state
to any other state, but also it needs to {visit some states in a
	specific temporal order.} At the beginning of each episode, the agent
is initialized in an arbitrary location in an arbitrary room. The
agent has four possible move actions, $\mathcal{A}
= \{North, South, East, West\}$, on each time step. The agent receives
$r=+10$ reward for reaching the key and $r=+40$ if it moves to the
lock while carrying the key (i.e., any time after visiting the key
location during the same episode). Bumping into a wall boundary is
punished with a reward of $r=-2$. There is no reward for just
exploring the space. Learning in this environment with sparse delayed
feedback is challenging for a reinforcement learning agent. To
successfully solve the task, the agent should represent knowledge at
multiple levels of spatial and temporal abstractions. The agent should
also learn to explore the environment efficiently.

\begin{figure*}[hbt!] 
	\centering
	\begin{tabular}{cc}
		\includegraphics[width=0.47\textwidth]{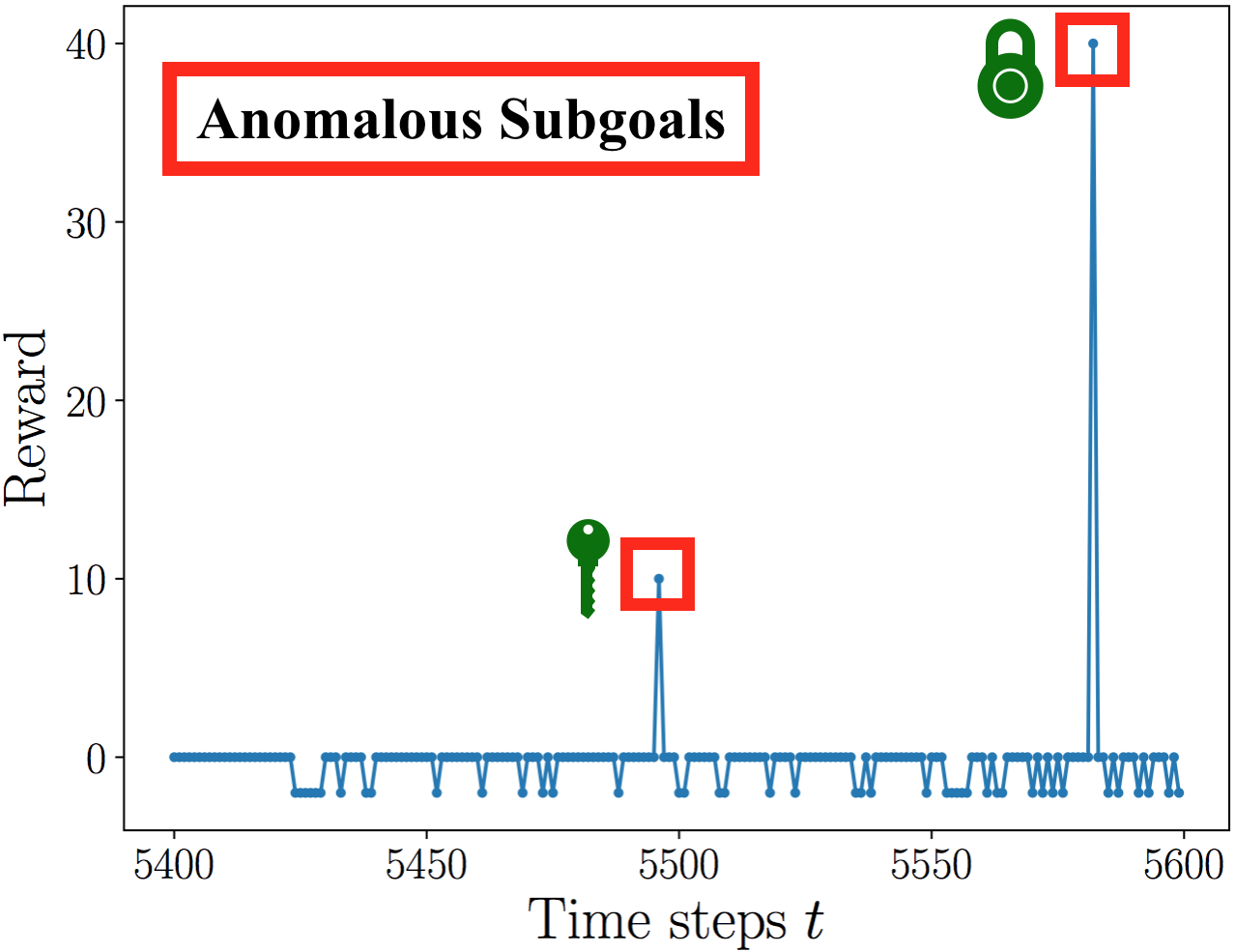} & \includegraphics[width=0.47\textwidth]{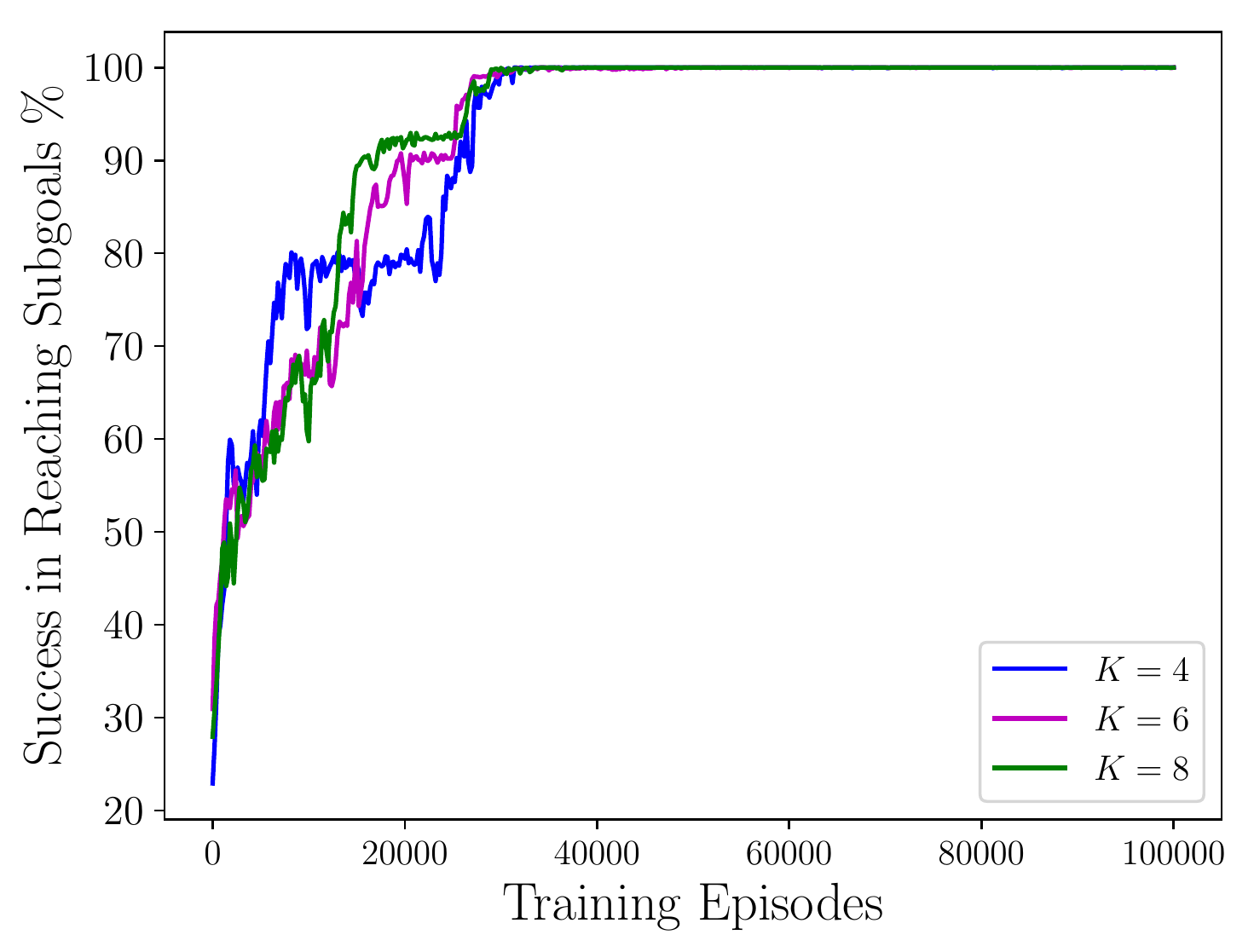}\\
		(a) & (b)  \vspace{0.5cm}\\
		\includegraphics[width=0.47\textwidth]{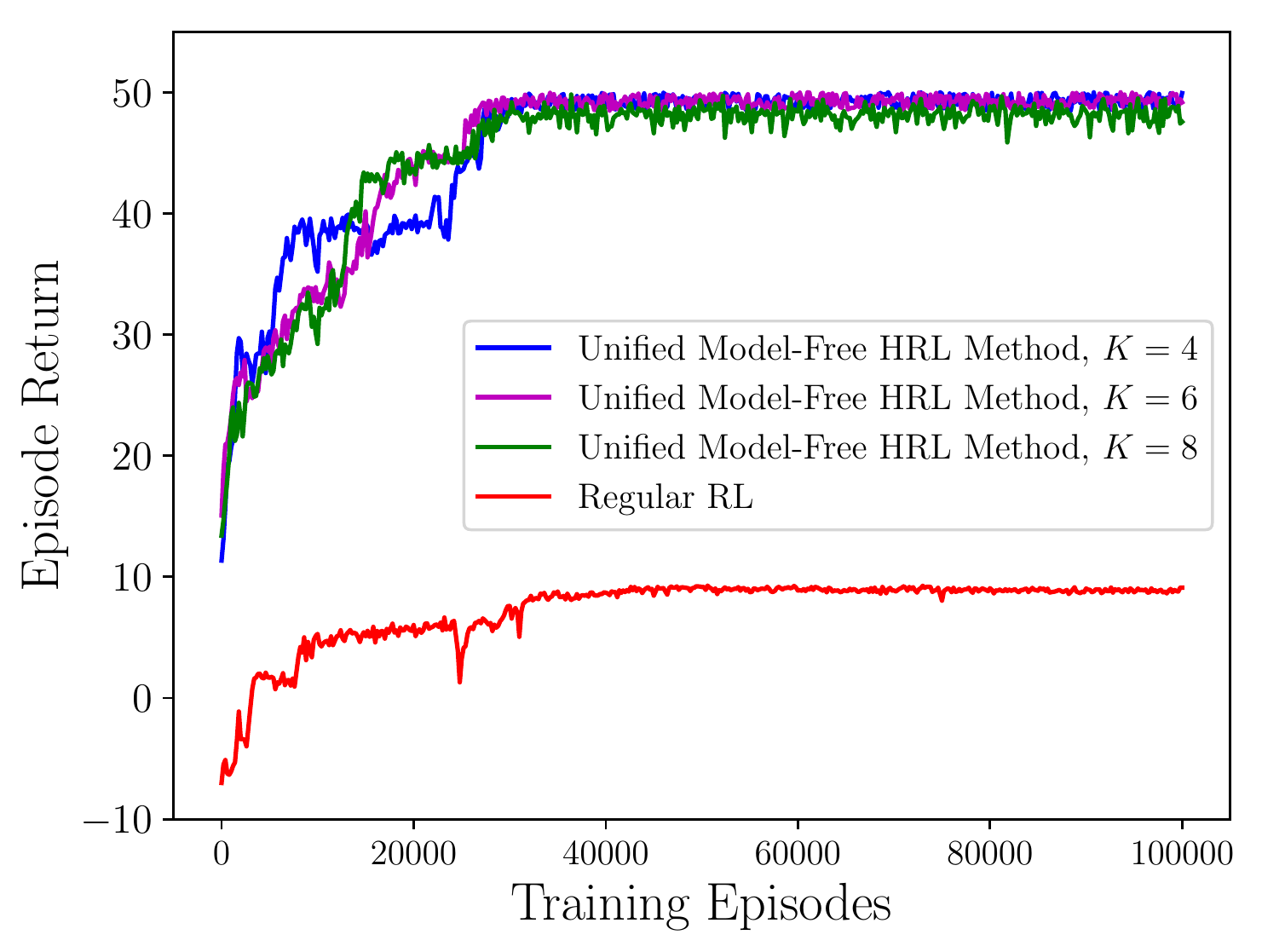} &
		\includegraphics[width=0.47\textwidth]{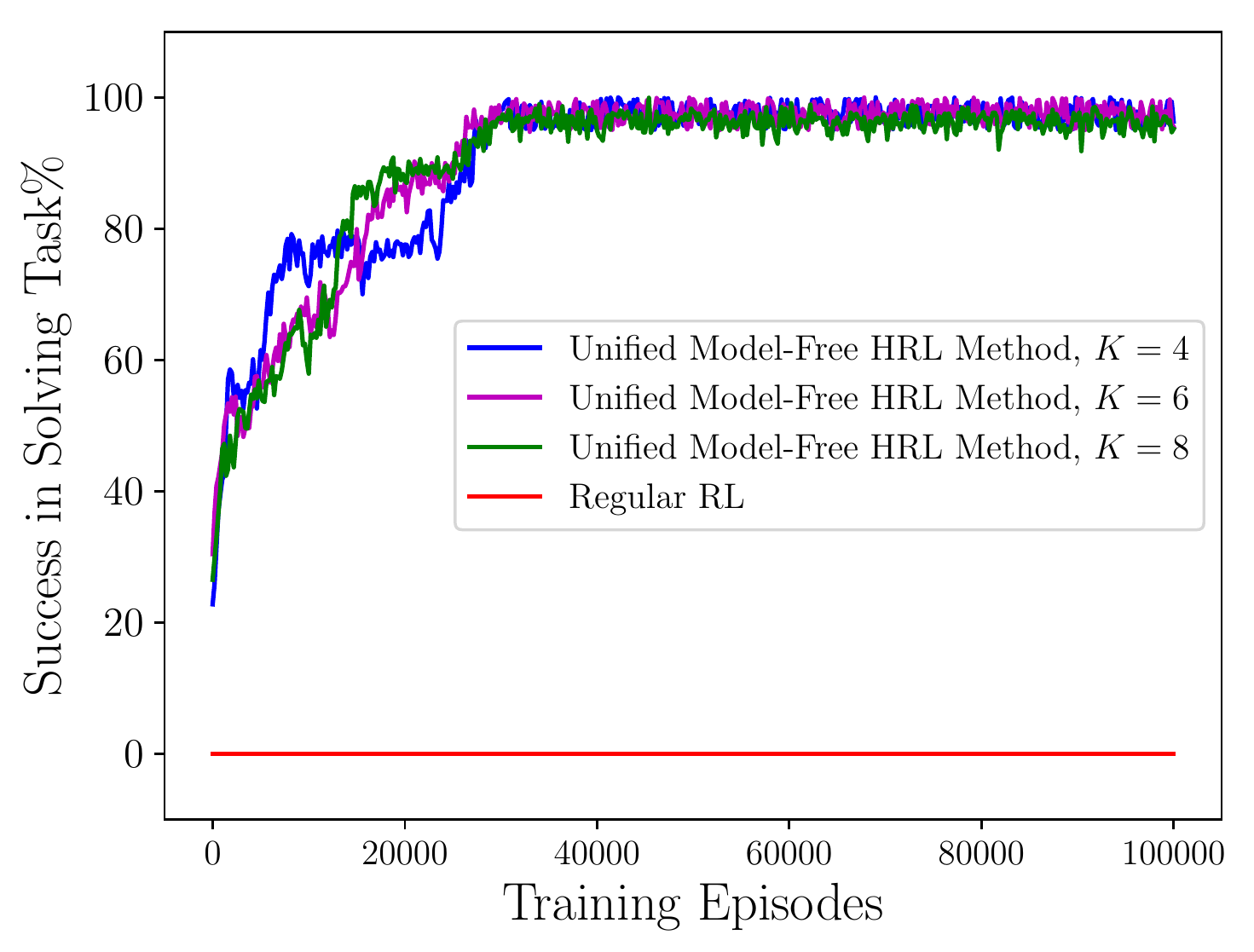} \\
		(c) & (d)  
	\end{tabular}
	\caption{(a) Reward over an episode, with anomalous points corresponding to the key ($r=+10$) and the lock ($r=+40$). (b) The average success of the controller in reaching subgoals over 200 consecutive episodes. (c) The average episode return. (d) The average success rate for solving the 4-room task.}
	\label{fig:rooms-results}
\end{figure*}

We first examined the unsupervised subgoal discovery algorithm over the course of a random walk. The agent was allowed to explore the \emph{4-room} environment for $100$ episodes. Each episode ended either when the task was completed or after reaching a maximum time step limit
of $200$. The agent's experiences, $e =(s,a,r,s')$, were collected in
an experience memory, $\mathcal{D}$. The stream of external rewards
for each transition was used to detect \emph{anomalous} subgoals
(Figure \ref{fig:rooms-results}(a)). We applied a heuristic anomaly detection method for the streaming rewards that was able to differentiate
between the rare large {positive} rewards and the regular small
ones. These peaks, as shown in Figure \ref{fig:rooms-results}(a), corresponded to the experiences in which the key was reached ($r=+10$) or the experience of reaching the lock after obtaining the key.

We also applied a $K$-means clustering algorithm to the experience
memory. (See Algorithm \ref{Algo:unified-model-free-hrl}.) The
centroids of the $K$-means clusters (with $K=4$) are plotted in
Figure \ref{fig:rooms}(b). The clusters, along with the anomalous states, were collected into the set of subgoals. By choosing $K=4$ for the number of clusters (or centroids), the discovered centroids roughly correspond to the centers of the rooms, and the clusters correspond to the rooms. But, the choice of $K=4$ comes from our knowledge of the spatial structure of this environment. Here, we show that other choices for $K$ leads themselves to different, but still useful, clusterings of the state space. Indeed, all we expect from the clustering algorithm in unsupervised subgoal discovery is to divide the state space into clusters of states with roughly similar values, and the choice of $K$ is not crucial. Also, any good spatial clustering method would work equally well. For example, with the number of clusters, $K=6$, we saw a clustering in which, two of the four rooms were divided into two clusters (see Figure \ref{fig:rooms}(c)). We repeated this experiment for $K=8$, where we saw equally useful clusters, with two room containing two cluster centroids, one room containing three clusters, and one room only with one cluster (see Figure \ref{fig:rooms}(d)). The K-means algorithm in our unsupervised subgoal discovery can be incremental, using the previous centroids as initial points for the next iteration. Therefore, the configuration of clusters were different over training, but the results of clustering were useful regardless. 

In summary, our unified model-free HRL framework (Algorithm \ref{Algo:unified-model-free-hrl}) does not rely on a particular choice of $K$. To prove this claim, we trained the agent in the unified model-free HRL framework for different numbers of clusters, $K=4$, $K=6$, and $K=8$.

\begin{figure}[htb!]
	\centering
	\includegraphics[width=0.7\textwidth]{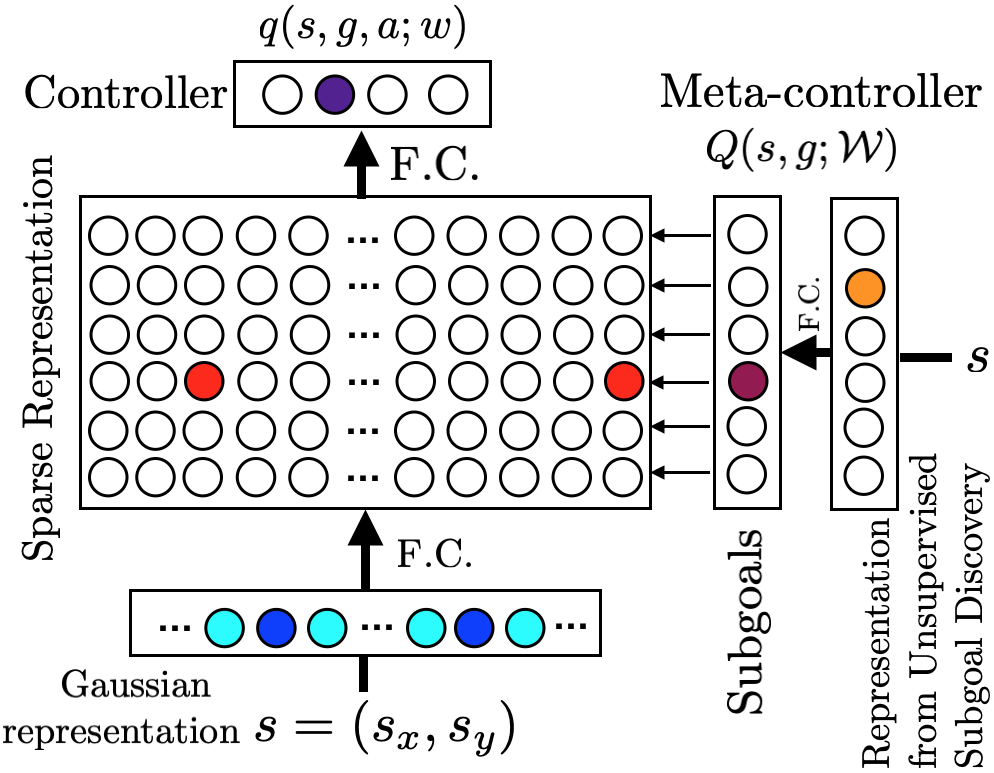} 
	\caption{Integrated meta-controller and controller network architecture.}
	\label{fig:unified-rooms-network}
\end{figure}

In these simulations, learning occurred in one unified phase consisting of 100,000 episodes. The meta-controller and the controller, and unsupervised subgoal discovery, were trained all together. See Algorithm \ref{Algo:unified-model-free-hrl}. Value function approximators were implemented as multi-layer artificial neural networks as shown in Figure \ref{fig:unified-rooms-network}. The controller network,
$q(s,g,a;w)$, took the state, $s$, and the goal, $g$, as
inputs. States were presented to the network as Cartesian coordinates,
with separate pools of inputs for each of the two
dimensions. The subgoal was initially chosen randomly from the set of discovered subgoals, resulting from unsupervised subgoal discovery during early random walks of the agent. 
The meta-controller value function receives a one-hot encoding of the current state, computed by converting the current state to the index of the corresponding subgoal. The meta-controller outputs a one-hot encoding of the best subgoal. The controller receives a Gaussian-blurred representation of current state variables (Cartesian coordinates) gated by the current subgoal, and it produces a sparse conjunctive encoding over hidden units using a $k$-Winners-Take-All mechanism, akin to lateral inhibition in cortex \citep{Rafati-Noelle:2015:CSC,OReillyRC:2001:CECN}. This is then mapped onto the controller value function output for each possible action. Most previously published subgoal discovery methods focus on finding the doorways (funnel type subgoals) \citep{Goel:2003:Subgoal,Simsek:2005:subgoal}. With $K=4$, the doorways can be discovered as boundaries between adjacent clusters. Note that our method is not strongly task dependent, so the choice of $K$ is not crucial to the learning of meaningful representations. The results of clustering for different values of $K$ are shown in Figure \ref{fig:rooms} (b-d). 

When a centroid was selected as a subgoal, if the agent entered any state in the corresponding cluster, the subgoal was considered attained. Thus, the controller essentially learned how to navigate from any location to
any state cluster and also to any of the anomalous
subgoals (key and door). The learning rate was $\alpha=0.001$, the
discount factor was $\gamma=0.99$, and the exploration rate was set to
$\epsilon_1=\epsilon_2=0.2$. The average success rate of the controller in achieving subgoals is shown in Figure \ref{fig:rooms-results}(b).

The average return, over 200 consecutive episodes, is shown in
Figure \ref{fig:rooms-results}(c). The agent very quickly converged on the
optimal policies and collected the maximum reward ($+50$). The high
exploration rate of $0.2$ caused high stochasticity, but the
meta-controller and controller could robustly solve the task on more
than $90\%$ of the episodes very early in training. After about 40,000
episodes, the success rate was $100\%$, as shown in
Figure \ref{fig:rooms-results}(d). There was no significant difference in the results of learning for different choice of the number of clusters $K$.

We compared the learning efficiency of our unified HRL method with the
performance resulting from training a {value approximation}
network with a regular, {non-hierarchical}, RL algorithm, TD
SARSA \citep{RL-Book:Sutton:Barto:2017}. The function approximator that
we used for $Q(s,a;w)$ matched that of the controller, equating for
computational resources, and we used the same values for the training
hyper-parameters. The regular RL agent could only reach the key before
becoming stuck in that region, due to the high local reward. Despite
the very high exploration rate used, the regular RL agent was not
motivated to explore the rest of the state space to reach the lock and
solve the task. Results are shown in Figure \ref{fig:rooms}(c) and (d)
(red lines).

It is worth noting that this task involves a \emph{partially observable Markov decision process} (POMDP), because information about whether or not the agent has the key is not visible in the state. This hidden state information poses a serious problem for standard RL algorithms, but our HRL agent was able to overcome this obstacle. Through meta-controller learning, the hidden information became implicit in the selected subgoal, with the meta-controller changing the current subgoal once the key is obtained. In this way, our HRL framework is able to succeed in task environments that are effectively outside of the scope of standard RL approaches.

\subsection{Montezuma's Revenge}
\begin{figure*}[hbt!] 
	\centering
	\begin{tabular}{cccc}
		\includegraphics[width=0.2\textwidth]{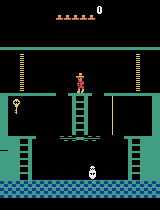}&
		\includegraphics[width=0.22\textwidth]{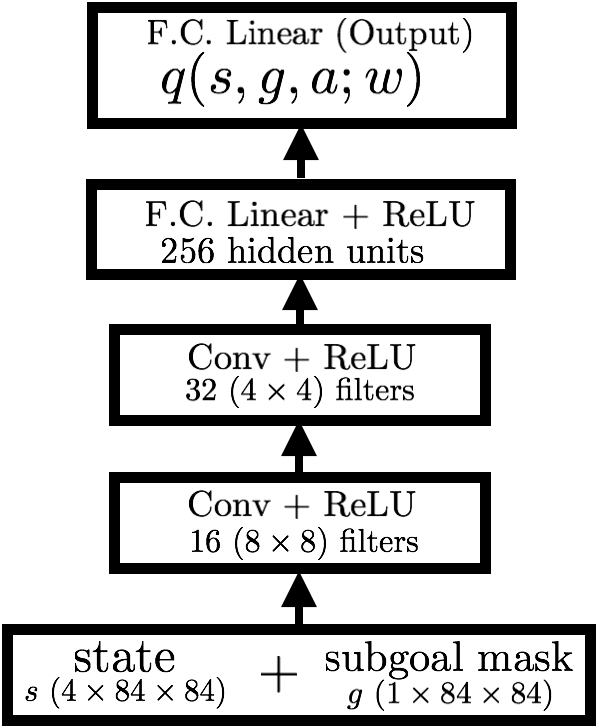} &
		\includegraphics[width=0.16\textwidth]{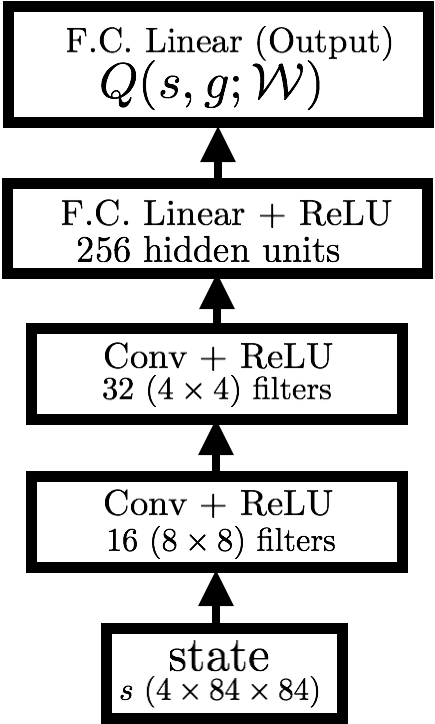} &
		\includegraphics[width=0.2\textwidth]{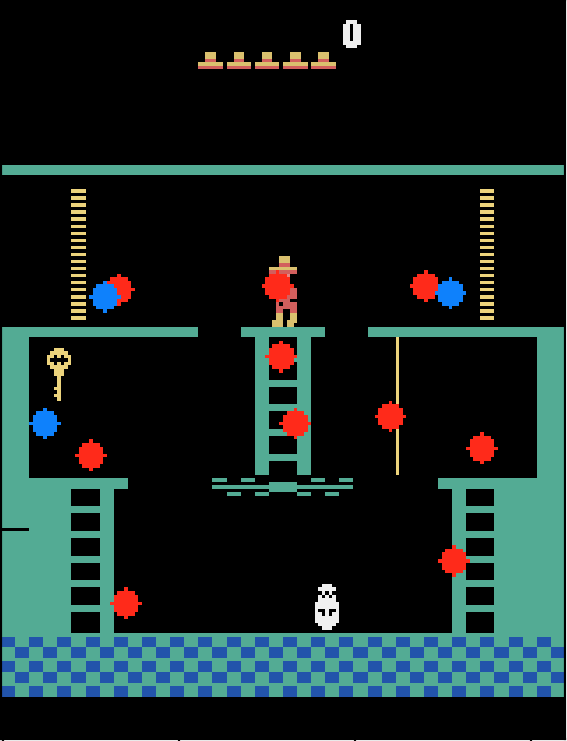}\\
		(a) & (b) & (c) & (d) \vspace{1cm} \\
		
		\multicolumn{2}{c}{\includegraphics[width=0.45\textwidth]{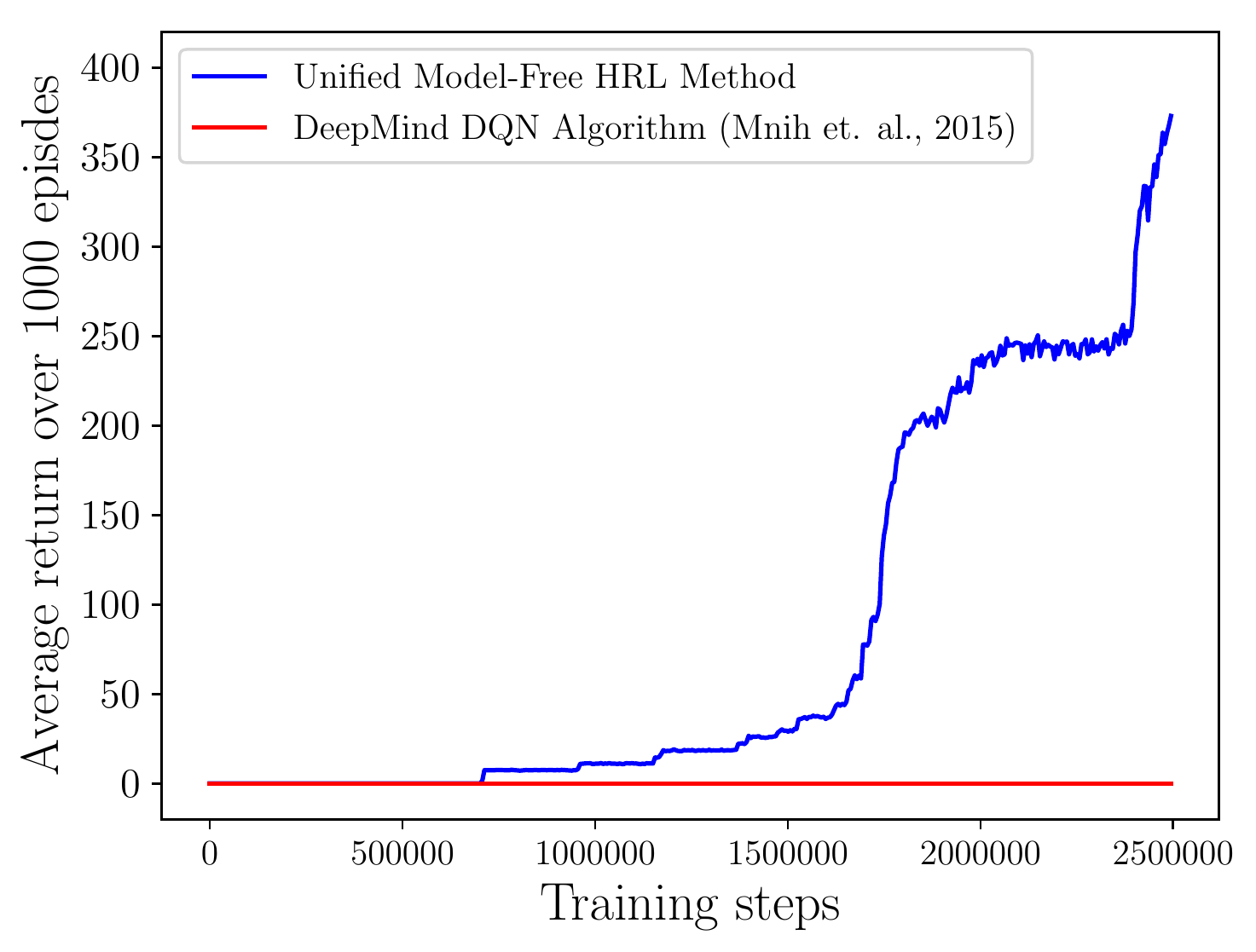} } & \multicolumn{2}{c}{\includegraphics[width=0.45\textwidth]{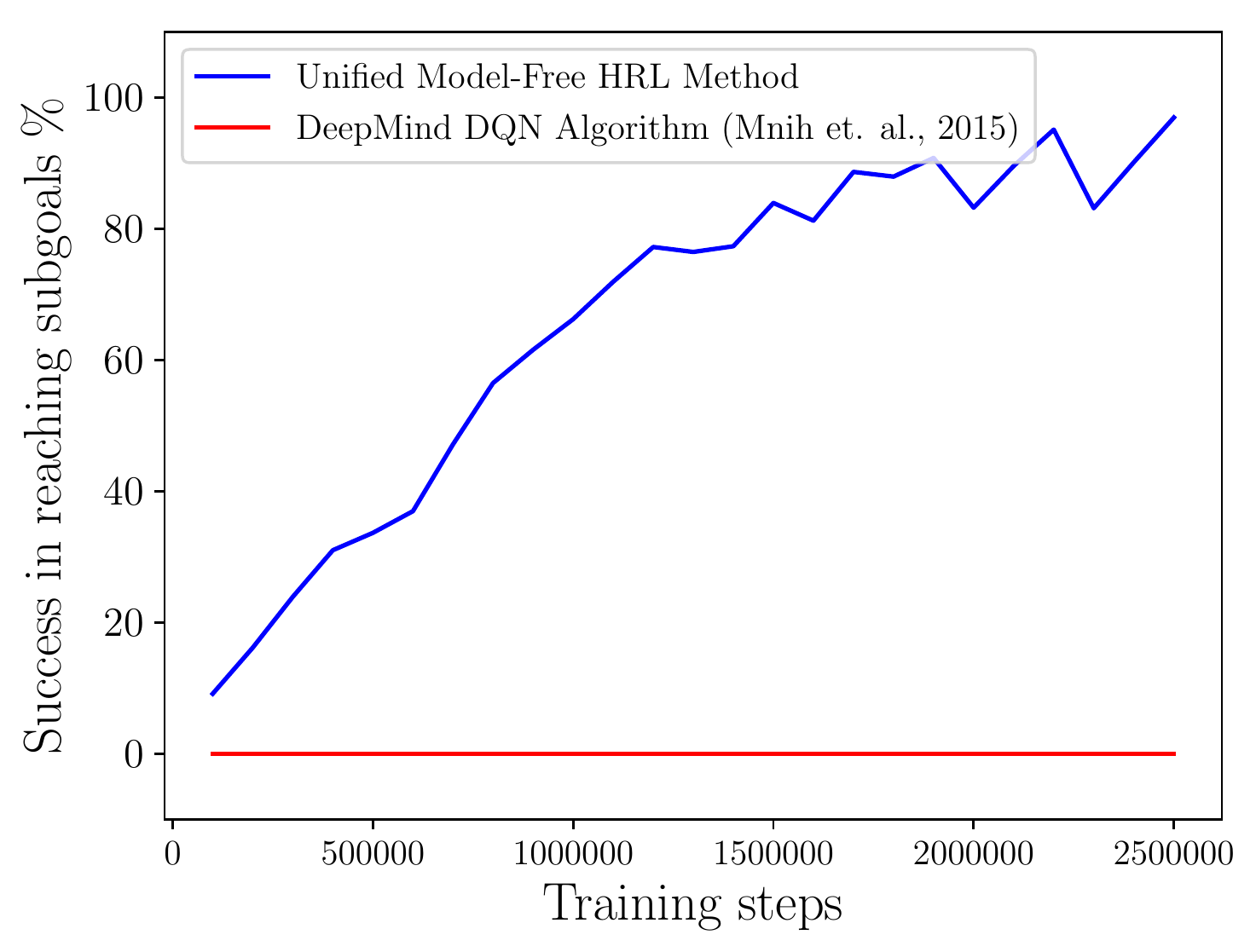} }\\
		\multicolumn{2}{c}{(e)} & \multicolumn{2}{c}{(f)}
	\end{tabular}
	\caption{(a) A sample screen from the ATARI 2600 game \emph{Montezuma's Revenge}. (b) The CNN architecture for the controller's value function. (c)  The CNN architecture for the meta-controller's value function. (d) The results of the unsupervised subgoal discovery algorithm. The blue circles are the discovered anomalous subgoals and the red ones are the centroid subgoals. (e) The average of return over 1000 episodes during the second phase of the learning. (f) The success of the controller in reaching to the subgoals during the second phase of learning.}
	\label{fig:montezuma}
\end{figure*}

We applied our HRL approach to the first room of the
game \emph{Montezuma's Revenge}. (See Figure \ref{fig:montezuma}(a).)
The game is well-known as a challenge for RL agents because it
requires solving many subtasks while avoiding traps. Having only
sparse delayed reward feedback to drive learning makes this RL problem
extremely difficult. The agent should learn to navigate the \emph{man}
in red to the \emph{key} by: (1) climbing the \emph{middle ladder} (2)
climbing the \emph{bottom right ladder} (3) climbing the \emph{bottom
	left ladder} (4) moving to the key. After picking up the key
($r=+100$), the agent should return back, reversing the previous
action sequence, and attempt to reach the \emph{door} ($r=+300$) and
exit the room. The moving skull at the bottom of the screen, which
ends an episode upon contact, makes obtaining the key extremely
difficult. {The episode also ends unsuccessfully if the man falls
	off of a platform.}

DeepMind's Deep Q-Learning (DQN)
algorithm \citep{DeepMind:Nature:2015}, which surpassed human 
performance on many ATARI 2600 games, failed to learn this game since
the agent did not reach any rewarding state during exploration.

In this problem, the agent requires the skills arising from intrinsic
motivation learning in order to explore the environment in a more
efficient way \citep{Kulkarni:2016:Meta-Controller}. Our HRL approach
supports the learning of such skills. The meta-controller
and the controller were trained in two phases. 

In Phase I (pretraining), the controller was trained to move the man from any location in the given frame, $s$, to any other location specified in a subgoal frame, $g$. An initial set of ``interesting'' subgoal locations were identified using a custom edge detection algorithm, avoiding empty
regions as subgoals. Unsupervised object detection using computer 
vision algorithms can be challenging
\citep{Kulkarni:2016:Meta-Controller,Fragkiadaki:2015:CVPR:ObjectDetection}. We made the simplifying assumption that, in many games, edges were
suggestive of objects, and the locations of objects made for good
initial subgoals. These locations were used {in Phase I of
	training} to train the controller through intrinsic
motivation using Algorithm \ref{Algo:intrinsic-motivation}. {Note that edge detection was only performed to
	identify Phase I subgoals. Specifically, it was \emph{not} used to
	change or augment the state representation in any way.}

We used a variant of the DQN deep Convolutional Neural Network (CNN)
architecture for approximation of the
controller's value function, $q(s,g,a;w)$ (see Figure \ref{fig:montezuma} (b)). The input to the controller
network consisted of four consecutive frames of size $84 \times 84$,
encoding the state, $s$, and an additional frame binary mask encoding
the subgoal, $g$. The concatenated state and subgoal frames were
passed to the network, and the controller then selected one of 18
different joystick actions based on a policy derived from
$q(s,g,a;w)$.

During intrinsic motivation learning, the recent experiences were
saved in an experience memory, $\mathcal{D}$, with a size of
$10^6$. In order to support comparison to previously published
results, we used the same learning parameters as DeepMind's
DQN \citep{DeepMind:Nature:2015}. Specifically, the learning rate, $\alpha$, was set to to be $0.00025$, with a discount rate of
$\gamma=0.99$. During Phase I learning, we trained the network for a
total of $2.5 \times 10^6$ time steps. The exploration probability
parameter, $\epsilon_1$, decreased from $1.0$ to $0.1$ in the first million
steps and remained fixed after that. 

After every $100,000$ time steps, we applied our unsupervised subgoal discovery method to the contents of the experience memory in order to find new subgoals, both anomalies and centroids, using $K$-means clustering with $K=10$. As shown in
Figure \ref{fig:montezuma}(d), the unsupervised learning algorithm
managed to discover the location of the key and the doors in this
way. It also identified useful objects such as ladders, platforms, and
the rope. These subgoals were used to train the meta-controller, and controller.

In Phase II, we trained the meta-controller, the
controller, and unsupervised subgoal discovery jointly together using Algorithm \ref{Algo:unified-model-free-hrl}. We reset the exploration rates, $\epsilon_1$, and $\epsilon_2$ to 1. The exploration probability
parameters decreased from $1.0$ to $0.1$ in the first million
steps and remained fixed after that. We ran the unsupervised subgoal discovery method every $100,000$ time steps to update the centroids of the clusters. We used an architecture based on the DQN CNN \citep{DeepMind:Nature:2015}, as shown in Figure \ref{fig:montezuma}(c), for the meta-controller's value 
function, $Q(s,g;\mathcal{W})$. All the rewarding (anomalous) subgoals were discovered in the Phase I. We used the non-overlapping discovered subgoals, which resulted in a set of 11 subgoals, $\mathcal{G}$. At the beginning of each episode, the meta-controller assigned a subgoal, $g \in \mathcal{G}$, based on an epsilon-greedy policy derived from $Q(s,g;\mathcal{W})$. The controller then attempted to reach these subgoals. The controller's experience memory, $\mathcal{D}_1$, had a
size of $10^6$, and the size of the meta-controller's experience
memory, $\mathcal{D}_2$, was $5\times 10^4$. 

The cumulative rewards for the game episodes are shown in Figure \ref{fig:montezuma}(e). After about 1.8 million time steps, the controller managed to reach the key subgoal more frequently. The average success of the intrinsic motivation learning over $100,000$ consecutive episodes is depicted in Figure \ref{fig:montezuma}(f). After about $2$ million learning steps, the
meta-controller regularly chose the proper subgoals for collecting the
maximum reward ($+400$).  

\section{Neural Correlates of Model-Free HRL}
This work has been inspired, in part, by theories of reinforcement learning in the brain. These theories often involve interactions between the striatum and neocortex. The temporal difference learning algorithm, which is a model-free RL, account for the role of the midbrain dopaminergic system \citep{SchultzW:1993:Dopamine}. The actor-critic architectures for RL have drawn connections within the basal ganglia and cerebral cortex. The RL-based accounts have also addressed the learning processes for motor control, working memory, and habitual and goal-directed behaviors.

There is some evidence that temporal abstraction in HRL might map onto regions within the dorsolateral and orbital prefrontal cortex (PFC) \citep{Botvinick:2009:HRL}, allowing the PFC to provide hierarchical representations to the basal ganglia. 

More recent discoveries reveal a potential role for medial temporal lobe structures, including the hippocampus, in planning and spatial navigation \citep{Botvinick2014MBHRL}, utilizing a hierarchical representation of space \citep{Chalmers2016HRLHippo}. There are evidences that hippocampus serve in model-based and model-free HRL with both flexibility and computational efficiency \citep{Chalmers2016HRLHippo}. Perhaps, the most salient aspects of the hippocampus is the existence of \emph{place cells}. \citep{Strange2014PlaceCells}, which activate in particular regions of an environment. Place cells in the dorsal hippocampus represent small regions while those in the ventral hippocampus represent larger regions \citep{Chalmers2016HRLHippo}. The fact that the hippocampus learns representations at multiple scales of abstraction supports the idea that the hippocampus might be a major component of the subgoal discovery mechanism in the brain. For navigation in the 4-room task, we see that the clustering algorithm divides the state space into a few big regions (ventral hippocampus), and the anomaly detection algorithm detects much smaller rewarding regions (dorsal hippocampus).     

There are also studies of interactions between the hippocampus and the PFC that are directly related to our unsupervised subgoal discovery method. \cite{Preston:2013:PFC-Hippocampus} illustrated how novel memories (like \emph{anomalous} subgoals) could be reinforced into permanent storage. Additionally, their studies suggest how PFC may be important for finding new meaningful representations from memory replay of experiences. This phenomena is similar to our clustering of experience memory.

\section{Conclusions}
We have proposed and demonstrated a novel model-free HRL method for
subgoal discovery using unsupervised learning over a small memory of
the most recent experiences {(trajectories)} of the agent. When
combined with an intrinsic motivation learning mechanism, this method
learns subgoals and skills together, based on experiences in the
environment. Thus, we offer an HRL approach that does not require a
model of the environment, making it suitable for larger-scale
applications.

Our results show that the intrinsic motivation learning produces a good policy to explore the states space efficiently, which leads to successful subgoal discovery. Our unsupervised subgoal discovery mechanism is able to find the structure of the state space, and learns the spatial hierarchies, and the meta-controller learns the temporal hierarchies to choose subgoals in the correct order. 

We hypothesize that the hippocampus, in concert with the prefrontal cortex, is playing a major role in the subgoal discovery process by replaying the  memory of experiences, in order to find meaningful low dimensional representation of the state.


\end{document}